\documentclass[10pt,journal,compsoc]{IEEEtran}

\usepackage{amsmath,amsfonts}
\usepackage{algorithmic}
\usepackage{algorithm}
\usepackage{array}
\usepackage[caption=false,font=normalsize,labelfont=sf,textfont=sf]{subfig}
\usepackage{textcomp}
\usepackage{stfloats}
\usepackage{url}
\usepackage{verbatim}
\usepackage{graphicx}
\usepackage{cite}
\usepackage{tikz}
\usepackage{scalerel}
\usepackage[edges]{forest}

\definecolor{hidden-draw}{RGB}{0,51,102}
\definecolor{hidden-blue}{RGB}{194,232,247}
\definecolor{hidden-orange}{RGB}{243,202,120}
\definecolor{hidden-yellow}{RGB}{242,244,193}
\definecolor{tree-level-1}{RGB}{245,20,85}
\definecolor{tree-level-2}{RGB}{246,86,118}
\definecolor{tree-level-3}{RGB}{248,177,193}
\definecolor{tree-leaf}{RGB}{176,230,198}

\definecolor{Self}{RGB}{255,0,128}
\definecolor{Ensemble}{RGB}{0,127,255}
\definecolor{Iterative}{RGB}{153,51,255}

\definecolor{exemplar1}{RGB}{136,98,148}
\definecolor{exemplar2}{RGB}{148,210,242}
\definecolor{knowledge1}{RGB}{249,219,152}
\definecolor{knowledge2}{RGB}{255,245,220}

\usepackage{microtype}
\usepackage{latexsym}
\usepackage{soul}
\usepackage{amsmath}
\usepackage{booktabs}
\usepackage{bm}
\usepackage{hyperref}
\usepackage{cleveref}
\usepackage{graphicx}
\usepackage{tabularx}
\usepackage{soul}
\usepackage{xcolor}
\usepackage{todonotes}
\usepackage{multirow}
\usepackage{xpatch}
\usepackage{blindtext}
\usepackage{fdsymbol}
\usepackage{microtype}

\usepackage{adjustbox}
\usepackage{textcomp}
\usepackage{xparse}
\usepackage{enumitem}
\usepackage{makecell}
\usepackage{subcaption}
\usepackage{colortbl}

\usepackage{xparse}
\usepackage{stfloats}

\usepackage{pifont}%
\newcommand{\cmark}{{\color{normalgreen} \ding{51}}}%
\newcommand{\xmark}{{\color{darkred} \ding{55}}}%

\usepackage{amsmath}
\DeclareMathOperator*{\argmax}{arg\,max}

\begin{document}

\title{From Pixels to Insights:\\ A Survey on Automatic Chart Understanding\\ in the Era of Large Foundation Models}

\author{Kung-Hsiang Huang,
        Hou Pong Chan,
        May Fung,
        Haoyi Qiu \\
        Mingyang Zhou,
        Shafiq Joty,
        Shih-Fu Chang,
        Heng Ji

\IEEEcompsocitemizethanks{

\IEEEcompsocthanksitem Kung-Hsiang Huang: Salesforce AI Research. kh.huang@salesforce.com.
\IEEEcompsocthanksitem 
Hou Pong Chan, May Fung, Heng Ji: University of Illinois Urbana-Champaign. \{hpchan, yifung2, hengji\}@illinois.edu.
\IEEEcompsocthanksitem Haoyi Qiu: University of California, Los Angeles. haoyiqiu@cs.ucla.edu.
\IEEEcompsocthanksitem Mingyang Zhou: Capital One. mingyang.zhou@capitalone.com.
\IEEEcompsocthanksitem Shafiq Joty: Salesforce AI Research \& NTU. sjoty@salesforce.com.
\IEEEcompsocthanksitem  Shih-Fu Chang: Columbia University. sc250@columbia.edu.

}
}

\markboth{TRANSACTIONS ON KNOWLEDGE AND DATA ENGINEERING (TKDE) 2024.}%
{Shell \MakeLowercase{\textit{et al.}}: A Sample Article Using IEEEtran.cls for IEEE Journals}

\maketitle
\newcolumntype{P}[1]{>{\centering\arraybackslash}p{#1}}

\newcommand{\samsum}[1]{\textsc{SAMSum}}
\newcommand{\dialogsum}[1]{\textsc{DialogSum}}
\newcommand{\mixandmatch}[1]{\textsc{MixAndMatch}}
\newcommand{\confit}[1]{\textsc{ConFiT}}
\newcommand{\ctrldiasumm}[1]{\textsc{CtrlDiaSumm}}
\newcommand{\cods}[1]{\textsc{CODS}}
\newcommand{\modelshort}[1]{\textsc{C2TFec}}
\newcommand{\modelshortda}[1]{\textsc{ZeroFEC-DA}}
\newcommand{\datashort}[1]{\textsc{Chocolate}}
\newcommand{\pgn}[1]{\textsc{PGN}}
\newcommand{\cliff}[1]{\textsc{CLIFF}}
\newcommand{\conseq}[1]{\textsc{ConSeq}}
\newcommand{\vescore}[1]{\textsc{ChartVE}}

\definecolor{c2}{RGB}{218,0,0}
\newcommand{\propaHighlight}[1]{{\color{c2} {#1}}}
\definecolor{lightblue}{RGB}{212, 235, 255}
\definecolor{lightorange}{RGB}{255, 204, 168}
\definecolor{lightyellow}{RGB}{255, 255, 168}
\definecolor{lightred}{RGB}{255, 168, 168}
\definecolor{darkred}{RGB}{234, 107, 102}
\definecolor{darkerblue}{RGB}{103, 136, 184}
\definecolor{lightgreen}{RGB}{144, 238, 144}

\definecolor{gold}{rgb}{0.83, 0.69, 0.22}
\definecolor{normalgreen}{rgb}{0.55, 0.8, 0.4}
\sethlcolor{lightblue}

\newcolumntype{Y}{>{\centering\arraybackslash}X}
\newcommand\hlc[2]{\sethlcolor{#1} \hl{#2}}

\NewDocumentCommand{\steeve}
{ mO{} }{\textcolor{gold}{\textsuperscript{\textit{Steeve}}\textsf{\textbf{\small[#1]}}}}
\NewDocumentCommand{\heng}
{ mO{} }{\textcolor{red}{\textsuperscript{\textit{Heng}}\textsf{\textbf{\small[#1]}}}}
\NewDocumentCommand{\ken}
{ mO{} }{\textcolor{purple}{\textsuperscript{\textit{Ken}}\textsf{\textbf{\small[#1]}}}}

\NewDocumentCommand{\mingyang}
{ mO{} }{\textcolor{blue}{\textsuperscript{\textit{Mingyang}}\textsf{\textbf{\small[#1]}}}}
\NewDocumentCommand{\yi}
{ mO{} }{\textcolor{lightgreen}{\textsuperscript{\textit{Yi}}\textsf{\textbf{\small[#1]}}}}
\NewDocumentCommand{\zhenhailong}
{ mO{} }{\textcolor{magenta}{\textsuperscript{\textit{Zhenhailong}}\textsf{\textbf{\small[#1]}}}}
\NewDocumentCommand{\haoyi}
{ mO{} }{\textcolor{violet}{\textsuperscript{\textit{Haoyi}}\textsf{\textbf{\small[#1]}}}}
\NewDocumentCommand{\chang}
{ mO{} }{\textcolor{orange}{\textsuperscript{\textit{Chang}}\textsf{\textbf{\small[#1]}}}}
\NewDocumentCommand{\shafiq}
{ mO{} }{\textcolor{yellow}{\textsuperscript{\textit{Shafiq}}\textsf{\textbf{\small[#1]}}}}

\definecolor{gold}{rgb}{0.83, 0.69, 0.22}

\newcommand{\Steeve}[1]{{\color{orange}#1}}
\newcommand{\markred}[1]{{\color{darkred}#1}}

\begin{abstract}

Data visualization in the form of charts plays a pivotal role in data analysis, offering critical insights and aiding in informed decision-making. Automatic chart understanding has witnessed significant advancements with the rise of large foundation models in recent years. Foundation models, such as large language models, have revolutionized various natural language processing tasks and are increasingly being applied to chart understanding tasks. This survey paper provides a comprehensive overview of the recent developments, challenges, and future directions in chart understanding within the context of these foundation models. %
We review fundamental building blocks crucial for studying chart understanding tasks. %
Additionally, we explore various tasks %
and their evaluation metrics and sources of both charts and textual inputs. %
Various modeling strategies are then examined, encompassing both classification-based and generation-based approaches, along with tool augmentation techniques that enhance chart understanding performance. Furthermore, we discuss the state-of-the-art performance of each task and discuss how we can improve the performance. Challenges and future directions are addressed, highlighting the importance of several topics, such as domain-specific charts, lack of efforts in developing evaluation metrics, and agent-oriented settings.
This survey paper aims to provide %
valuable insights and directions for future research in chart understanding leveraging large foundation models. 
\end{abstract}


\section{Introduction}

\textbf{The Significance of Chart Understanding in Information Communication}\quad In our contemporary world of multimedia information, where the volume and complexity of data continue to burgeon, the role of charts stands paramount in facilitating effective communication of factual information, conveying insights, and informing decision making. Across diverse fields spanning academia, scientific research, digital media, and business realms, charts serve as indispensable tools for translating raw data into comprehensible visual narratives. Their ability to encapsulate complex datasets in a concise and intuitive format empowers decision-makers to grasp key insights swiftly, aiding informed reasoning and strategic planning. Recognizing this pivotal role of charts in modern information dissemination, there has been a sustained interest within the computational community, as evidenced by a plethora of research in \textbf{automatic chart understanding}. %
In particular, works on \textit{chart question answering} \cite{kafle2018dvqa, ebrahimikahou2018figureqa, masry-etal-2022-chartqa, methani2020plotqa, kantharaj-etal-2022-opencqa}, \textit{chart captioning} \cite{hsu-etal-2021-scicap-generating, hoque2022chart, kantharaj-etal-2022-chart, tang-etal-2023-vistext}, \textit{chart-to-table conversion} \cite{liu-etal-2023-deplot, do2023llms}, \textit{chart fact-checking} \cite{akhtar2023chartcheck, akhtar-etal-2023-reading}, \textit{chart caption factual error correction} \cite{huang2023lvlms} %
have laid foundational frameworks for exploring the intricacies of chart semantics in chart understanding technologies.

\textbf{Challenges and Opportunities in Chart Understanding Amidst the Era of Large Foundation Models} \quad Traditional chart understanding work \cite{10192564,hoque2022chart} focuses on finetuning methods that generally experience limitations with respect to domain portability and reasoning robustness. Excitingly, the advent of large vision-language foundation models (e.g., GPT-4V \cite{openai2023gpt4v}, LLaVA \cite{liu2023improvedllava}) has spawned a paradigm shift in automated reasoning capabilities, catalyzing unprecedented advances across various multimedia cognitive tasks -- including progress in strong zero/few-shot reasoning capabilities through text-based prompting. Despite the impressive improvements, the domain of chart understanding still presents major challenges. Charts present a unique set of obstacles owing to their rich visual representations and nuanced semantics, delivered through complex and creative organization of information manifested in text, numerical data, plots, and graphics objects. From bar charts and line graphs to pie charts and scatter plots, each chart type employs a distinct visual syntax to convey data relationships, requiring sophisticated interpretative mechanisms beyond mere pixel-level pattern recognition. Charts serve as a means to uncover important insights, such as emerging trends, outliers that challenge assumptions, and relationships among variables that are not immediately apparent from raw data in tabular form.
They enable comparative analyses across data points, providing a visual platform for summarizing varied entities or time periods clearly.
Furthermore, the intrinsic diversity of underlying datasets, ranging from simple numerical relations to intricate multidimensional entities, adds another layer of complexity to the chart understanding task. Despite these challenges, automated chart understanding offers a gateway to unlock actionable insights buried within the pixels of visual narratives. By harnessing the capabilities of large foundation models, chart understanding demonstrates improved potential in bridging the gap between raw visual data and meaningful insights, thereby enabling high-level information extraction, logical reasoning, and decision-making. %

Several studies have surveyed the landscape of chart understanding research, yet these surveys often show limitations in covering comprehensive scopes or reviewing information of sufficient specificity.
Some surveys do not cover modern datasets employed in chart understanding research, as well as the most contemporary modeling approaches, such as those involving pre-trained vision-language models and large foundation models \cite{farahani2023survey,hoque2022survey}. Conversely, other surveys concentrate predominantly on the visualization aspect (i.e. the transformation of data into charts), thereby missing the nuanced task of chart interpretation \cite{wu2021ai4vis, shen2022towards, yang2023foundation, he2024leveraging}. This survey paper aims to bridge these gaps. \looseness=-1

 We first define automatic chart understanding and the fundamental building blocks to the problem formulation in \Cref{Sec:2}. We discuss the multifaceted nature of chart understanding, encompassing tasks from interpreting chart visuals to analyzing underlying data. We also illustrate how chart understanding overlaps with related works in \textit{natural image understanding} \cite{pratt2020grounded}, \textit{table understanding} \cite{zhang-etal-2020-table}, and \textit{document understanding} \cite{kim2022donut}. We explain the essential modeling components for chart understanding, such as vision encoders, OCR modules, and text decoders, and their role in converting raw chart images and text queries into useful insights.%
 Then, in \Cref{Sec:3}, we examine the datasets driving chart understanding research and the metrics for model evaluation. This section analyzes the sources, diversity, and limitations of these datasets, providing insights into the current chart understanding data landscape. It also reviews various evaluation metrics, underlining the necessity for robust and nuanced assessment methods. With insights from these characteristics, we further provide trending popular modeling strategies for automatic chart understanding. \Cref{Sec:4} delves into the diverse modeling strategies in chart understanding, including adaptations from natural image understanding, vision-language pre-training, and foundation models like large language models (LLMs) and large vision-language models (LVLMs). In particular, we emphasize the impact of choices in vision encoders and text decoders on model effectiveness and discuss tool augmentation's role in chart understanding. We conclude this section by showcasing the state-of-the-art performance on different chart understanding tasks and how we can improve upon them. \looseness=-1

Finally, \Cref{Sec:5} addresses the challenges and future directions in chart understanding. We highlight the importance of domain-specific charts, the need for comprehensive evaluation metrics, and the potential for adversarial settings to enhance model robustness and versatility. %
This survey paper concludes by identifying key areas for future research, such as developing models for complex charts, refining evaluation metrics, and diversifying datasets. We not only offer an in-depth overview of the current state of chart understanding but also set the stage for future advancements in this exciting intersection of data visualization and machine learning. \looseness=-1

\section{Background} \label{Sec:2}

\subsection{What is Automatic Chart Understanding?}
Charts are graphical representations of data that are used to present complex patterns in data in a concise and visually appealing manner. Common types of charts include line charts, bar charts, area charts, pie charts, and scatter plots. 
Automatic chart understanding aims to enable machines to interpret charts and derive meaningful information such as patterns, trends, and relationships within the data presented in the charts. 
Automatic chart understanding techniques have various real-world applications, such as assisting data analysts in discovering patterns from charts, answering queries, and helping individuals with visual impairments to access the information in charts.

However, automatic chart understanding is challenging as it demands substantial perceptual and reasoning effort. 
First, charts encompass complex compositions of fine-grained graphical marks (e.g., lines, dots, bars, etc.) and scene text (e.g., axis labels, titles, legends, etc.). 
Therefore, \textit{perception} abilities are required to understand the layout and spatial relations among these elements to extract meaningful information from the chart. 
Moreover, to fulfill a specific task or query, models often need to be equipped with \textit{reasoning} capabilities. These range from basic comparison operations for interpreting the relative magnitude between data points to complex mathematical abilities, such as summation, on the values extracted from charts. 

\subsection{Related Tasks}
\subsubsection{Natural Image Understanding}
\label{subsec:natural_image_understanding}
Natural image understanding focuses on the interpretation and analysis of photographic images, which capture scenes from the real world. These images can include landscapes, people, objects, and everyday situations. The objective of this field is to enable machines to recognize and make sense of visual information in a manner that replicates human vision. Natural image understanding encompasses tasks such as image captioning \cite{vinyals2015show, qiu-etal-2023-gender} and visual question answering \cite{antol2015vqa,hudson2019gqa}. The core difference between natural image understanding and chart understanding lies in the type of visual content and the nature of the interpretation tasks involved. While natural image understanding deals with unstructured visual data that represents natural scenes, chart understanding focuses on structured visual representations of data, such as bar graphs, line charts, and pie charts. Charts are designed to communicate specific quantitative information and trends. This makes the interpretation task centered around data extraction, recognition of graphical elements, and comprehension of the data narrative conveyed by the chart.

One of the major challenges in natural image understanding is the variability and complexity of real-world scenes, including variations in lighting, viewpoints, and occlusions. In contrast, chart understanding entails challenges such as recognizing and differentiating among various chart types, extracting embedded textual and numerical data, and understanding the logical and quantitative relationships depicted. Furthermore, charts can contain a high level of abstraction and symbolic elements (e.g., colors representing different categories, lines indicating trends), necessitating not just visual perception but also mathematical or analytical reasoning to interpret the information accurately.

\subsubsection{Table Understanding}

Table understanding involves interpreting information presented in tabular data \cite{pasupat-liang-2015-compositional, parikh-etal-2020-totto, eisenschlos-etal-2020-understanding, liu-etal-2022-tapex}. This task aims to recognize the table's layout and structure, identify the relationships between its rows and columns, and comprehend the semantic context embedded within the table's content. In contrast to charts, which employ graphical elements to visually represent data, tables arrange this data in a meticulously organized format. They utilize rows and columns to systematically display information, offering a direct view of the data's structure and interrelationships.

Although chart and table understanding both engage with structured data, chart understanding confronts a unique challenge absent in table understanding: the interpretation of visual encodings. Charts employ diverse visual encodings, including color, shape, size, and orientation, to illustrate and highlight various facets of the data. These visual encodings can convey nuances such as trends, distributions, and comparisons in a way that is more immediately perceptible and sometimes more intuitively understood than textual data. However, this also means that chart understanding requires an additional layer of visual perception and interpretation that is not necessary for tabular data. 

Furthermore, while it might be arguable that chart understanding tasks can be simplified into table understanding tasks by converting charts into underlying data tables. However, data tables are infrequently published alongside charts, and the process of automatically extracting data tables is fraught with challenges. The diversity of chart types and visual styles, coupled with the intricacies of reasoning about visual occlusion, makes automatic data table extraction prone to errors. This complexity underscores the distinct and intricate nature of chart understanding, highlighting its unique challenges compared to table understanding.

\subsubsection{Document Understanding}
Document understanding encompasses the comprehensive analysis of information contained within documents that are presented visually. These documents can vary widely in format and content, including but not limited to scanned paper documents \cite{huang2019icdar}, digital PDFs \cite{mathew2021docvqa}, presentation slides \cite{tanaka2023slidevqa}. The essence of visual document understanding lies in its ability to decode both the textual content and the layout structure of documents, understanding the integral relationship between various components such as headings, paragraphs, tables, figures, and footnotes.

The primary distinction between visual document understanding and chart understanding is the scope and nature of the visual content being analyzed. While chart understanding is specifically focused on interpreting graphical representations of data, such as bar charts, line graphs, and pie charts, visual document understanding encompasses a broader range of document types and aims to comprehend the full gamut of information a document may convey. This includes not only charts but also textual content and their spatial arrangements, which collectively deliver the document's message.

Additionally, charts in the context of chart understanding may originate from diverse sources beyond just traditional documents. They are often found on consensus websites, in digital media, and within academic literature, where they serve as standalone visualizations of data rather than components of a larger document. This distinction highlights the versatility and ubiquity of charts as tools for data representation and the necessity for models in chart understanding to focus specifically on the nuances of visual data interpretation, free from the constraints of document structure. \looseness=-1

\subsection{Problem Formulation and Fundamental Building Blocks}
We uniformly formulate existing chart understanding tasks as the following problem. 
Given a chart image $\mathbf{c}$ and a textual query $\mathbf{q}$, the goal is to predict a textual response $\mathbf{y}$. 
Both the textual query and textual response are sequences of tokens, 
i.e., $\mathbf{q}=[q_{1},\ldots,q_{l^{q}}]$ and $\mathbf{y}=[y_{1},\ldots,y_{l^{y}}]$, where $l^q$ and $l^y$ denote the length of $\mathbf{q}$ and $\mathbf{y}$, respectively.  
The response $y$ can be restricted to a fixed vocabulary, such as the candidate answers in a multiple-choice question or the class labels in the chart fact-checking task. 
Alternatively, the response can be open vocabulary in tasks such as open-ended question answering or chart captioning. %

A chart understanding model is typically built by combining the following computational modules: 

\noindent \textbf{Vision Encoder}:
It is often essential to extract visual features from the input chart image in order to understand the relations and spatial arrangements between the graphical elements and scene text in the chart. Therefore, conventional chart understanding models usually employ a vision encoder $\textsc{Encoder}_{\text{vis}}$ to map the input chart image to a visual feature matrix: $\mathbf{H}_{c}=\textsc{Encoder}_{\text{vis}}(\mathbf{c})$. %

\noindent \textbf{Chart-to-Table Conversion Module}:
The underlying data table $\mathbf{T}$ of an input chart provides a fully structured textual representation of its raw data. This can help language models better understand the information presented in the input chart. 
In real-world applications, the underlying data tables of charts may %
not readily accessible. Therefore, various chart understanding approaches employ a chart-to-table translation module to extract a data table $\hat{\mathbf{T}}$ from the input chart. %
An extracted table can also be linearized into a sequence of table tokens $[\hat{\mathbf{T}}_{1},\ldots, \hat{\mathbf{T}}_{l^{\hat{\mathbf{T}}}}]$, where $l^{\hat{\mathbf{T}}}$ denotes the number of elements in $\hat{\mathbf{T}}$. %

\noindent \textbf{OCR Module}: 
Recognizing the scene text in a chart image is an essential step for chart understanding. Many chart understanding methods apply an OCR (Optical Character Recognition) system to extract scene text from the input chart image: $\hat{\mathbf{s}} = \textsc{OCR}(\mathbf{c})$. 
The extracted scene text is a set of tokens $\hat{\mathbf{s}}=\{\hat{s}_{1},\ldots,\hat{s}_{l^{\hat{s}}}\}$, where $l^{\hat{s}}$ %
denotes the number of tokens in the extracted scene text. An OCR system also provides the positional metadata, known as bounding box, %
for each extracted token, including its top left coordinates, bottom right coordinates, width, and height. 

\noindent \textbf{Text Encoder}:
To understand the input textual query, a text encoder is often applied to map the input textual query to a query representation matrix: $\mathbf{H}_{q}=\textsc{Encoder}_{\text{txt}}(\mathbf{q})$. 
The text encoders in existing chart understanding models are commonly realized by a Transformer encoder~\cite{DBLP:conf/nips/Transformer17} or word embedding layer from a pre-trained language model~\cite{DBLP:conf/acl/BARt20}. Similarly, we can use a text encoder to encode an extracted table into a representation matrix $\mathbf{H}_{\hat{T}}$.

\noindent \textbf{Text Decoder}:
A text decoder sequentially generates a predicted textual response $\hat{\mathbf{y}}$ given an input context set $\mathcal{X}$. In existing chart understanding models, the input context set usually consists of the chart representation matrix, query representation matrix, and/or the representation matrix of the extracted table, e.g., $\mathcal{X}=(\mathbf{H}_{c},\mathbf{H}_{q},\mathbf{H}_{\hat{T}}$).
Specifically, a text decoder $\textsc{Decoder}_{\text{txt}}$ learns a conditional probability distribution over a variable length predicted response $\hat{\mathbf{y}}$:
$P(\hat{\mathbf{y}}|\mathcal{X})=\sum_{t=1}^{l^{\hat{y}}}P(\hat{y}_{t}|\hat{y}_{t-1}, \ldots, \hat{y}_{1}, \mathcal{X})$, where $\hat{y}_{1}$ denotes the $i$-th token in $\hat{\mathbf{y}}$ and $l^{\hat{y}}$ denotes the length of $\hat{\mathbf{y}}$.

Note that some language models, like GPT-3~\cite{brown2020language}, feature a decoder-only architecture, omitting a text encoder module. These models leverage a unified architecture that encodes texts in the decoder. Without loss of generality, we denote the decoder input that corresponds to the textual queries as \textit{encoding}, recognizing that this \textit{encoding} process resembles the encoder within encoder-decoder architectures.

Additionally, there are some variations in the way that input charts are encoded. For example, recently developed Large Vision-Language Models (LVLMs) like ChartLlama \cite{han2023chartllama} and ChartAssistant \cite{meng2024chartassisstant} employ an additional projection layer $\textsc{Projector}_{\text{vt}}$ to better align text and visual representations: $\mathbf{H}_{c}=\textsc{Projector}_{\text{vt}}(\textsc{Encoder}_{\text{vis}}(\mathbf{c}))$.

\begin{table}[t]
\centering
\caption{Notations used in this paper. }
\begin{tabular}{cp{.75\columnwidth}}
\toprule
\textbf{Notation} & \textbf{Description} \\ \hline
$\mathbf{c}$ & Input chart image \\
$\mathbf{H}_{c}$ & Representation matrix for the input chart image \\
$\mathbf{q}$ & Input textual query \\
$\mathbf{H}_{q}$ & Representation matrix for the input textual query \\
${q}_{i}$ & The $i$-th token of the input textual query. \\
$l^q$ & Number of tokens in the input textual query \\
$\hat{\mathbf{s}}$ & Scene text extracted from the input chart image \\
$\mathbf{y}$ & Gold standard textual response \\
$\hat{\mathbf{y}}$ & Predicted textual response \\
${\mathbf{T}}$ & Gold standard underlying data table of the input chart image \\
${\mathbf{T}}_{i}$ & The $i$-th token for the data table $\mathbf{T}$ after linearization \\
$\hat{\mathbf{T}}$ & Data table extracted from the input chart image by a system \\
$\textsc{Encoder}_{\text{vis}}$ & Vision encoder \\
$\textsc{Encoder}_{\text{txt}}$ & Text encoder \\
$\textsc{Decoder}_{\text{txt}}$ & Text decoder \\
$\textsc{Projector}_{\text{vt}}$ & Vision-language decoder \\
$\mathcal{X}$ & Input context set to the text decoder \\
\bottomrule
\end{tabular}
\vspace{-2mm}
\label{tab:notation}
\end{table}

\section{Tasks and Datasets} \label{Sec:3}

\begin{table*}[t]
    \small
    \centering
    \caption{Summary of chart understanding datasets. We define the following notions for tasks: \textit{FQA}: Factoid Question Answering, \textit{OQA}: Open-domain Question Answering, \textit{CAP}: Captioning, \textit{C2T}: Chart-to-Table, \textit{FC}: Fact-checking, \textit{FEC}: Factual Error Correction, \textit{FID}: Factual Inconsistency Detection. \cmark~/\xmark~ indicates the dataset consists of both types of data or charts.}%
    \begin{adjustbox}{max width=0.98\textwidth}
    {
    \begin{tabular}{llllrcc}
        \toprule
        \textbf{Dataset} & \textbf{Tasks} & \textbf{Source} & \textbf{Domain} & \textbf{\# Instances} & \textbf{Real-world Data} & \textbf{Real-world Charts} \\
        \midrule
        FigureQA \cite{ebrahimikahou2018figureqa} & FQA & Synthetic & General & 2M &\xmark~ & \xmark~\\
        DVQA \cite{kafle2018dvqa} & FQA & Synthetic & General & 3M & \xmark~ & \xmark~\\
        LEAF-QA \cite{chaudhry2020leaf}  & FQA & World Development Indicators et al. & General & 2M & \cmark~ & \xmark~\\
        LEAF-QA++ \cite{singh-shekhar-2020-stl} & FQA & World Development Indicators et al. & General & 2M & \cmark~ & \xmark~\\
        PlotQA \cite{methani2020plotqa} & FQA/C2T & World Bank Open Data et al.& General & 29M & \cmark~ & \xmark~\\
        ChartQA \cite{masry-etal-2022-chartqa} & FQA/C2T & Statista/Pew & General & 10K & \cmark~ & \cmark~\\
        MapQA \cite{chang2022mapqa} & FQA & Kaiser Family Foundation & General & 796K  & \cmark~ & \cmark~\\ %
        PaperQA \cite{lu2023mathvista} & FQA & Scientific Papers & Science & 107  & \cmark~ & \cmark~\\ %
        SciGraphQA \cite{li2023scigraphqa} & FQA & Arxiv & Science &  296K & \cmark~ & \cmark~\\
        MMC-Bench \cite{liu2023mmc} & FQA & Statista et al. & General & 2K & \cmark~ & \cmark~ \\
        ChartBench \cite{xu2023chartbench} & FQA & Kaggle & General &  17K & \cmark~ & \xmark~       \\
        ArXivQA \cite{li2024multimodal} & FQA & ArXiv & Science & 100K & \cmark~ & \cmark~ \\
        
        \midrule
        OpenCQA \cite{kantharaj-etal-2022-opencqa} & OQA & Pew & General & 8K  & \cmark~ & \cmark~\\ %
        \midrule
        FigCAP \cite{chen2019figure} & CAP & Synthetic & General & 101K & \xmark~ & \xmark~ \\
        Chart2text \cite{obeid-hoque-2020-chart} & CAP & Statista & General & 8K & \cmark~ & \cmark~\\ %
        SciCap \cite{hsu-etal-2021-scicap-generating} & CAP & ArXiv & Science & 476K & \cmark~ & \cmark~ \\
        LineCap \cite{mahinpei2022linecap} & CAP & ArXiv & Science & 4K & \cmark~ & \cmark~\\
        ChaTa+ \cite{seweryn2021will} & CAP &  ArXiv/World Health Organization & Science & 2K & \cmark~ & \cmark~ \\
        Chart-to-Text \cite{kantharaj-etal-2022-chart} & CAP/C2T  & Statista/Pew & General & 44K  & \cmark~ & \cmark~\\ %
        ChartSumm \cite{rahman2023chartsumm} & CAP & Statista/Knoema   & General     &   84K & \cmark~ & \cmark~ \\
        VisText \cite{tang-etal-2023-vistext} & CAP/C2T  & Statista & General & 12K  & \cmark~ & \xmark~\\ %
        FigCaps-HF \cite{singh2023figcaps} & CAP & ArXiv & Science & 134K & \cmark~ & \cmark~\\
        ArxivCap \cite{li2024multimodal} & CAP & ArXiv & Science & 4M & \cmark~ & \cmark~ \\
        \midrule
        ChartFC \cite{akhtar-etal-2023-reading} & FC & Wikipedia & General & 16K & \cmark~ & \xmark~ \\
        ChartCheck \cite{akhtar2023chartcheck} & FC & Wikimedia & General & 10K & \cmark~ & \cmark~\\ %
        \midrule
        Chocolate \cite{huang2023lvlms} & FEC/FID & Statista/Pew & General & 1K  & \cmark~ & \cmark~/\xmark~\\ %

        \bottomrule
    \end{tabular}
    }
    \end{adjustbox}
    \vspace{-5mm}
    
    \label{tab:dataset_summary}
    
\end{table*}

\Cref{tab:dataset_summary} shows an overview of the key characteristics of the available datasets and the corresponding tasks. %
In the following subsections, we analyze these datasets along four dimensions: the task, the source, and the type of plots.

\subsection{Tasks}

\begin{figure*}[t]
 \centering
 \includegraphics[width=\linewidth]{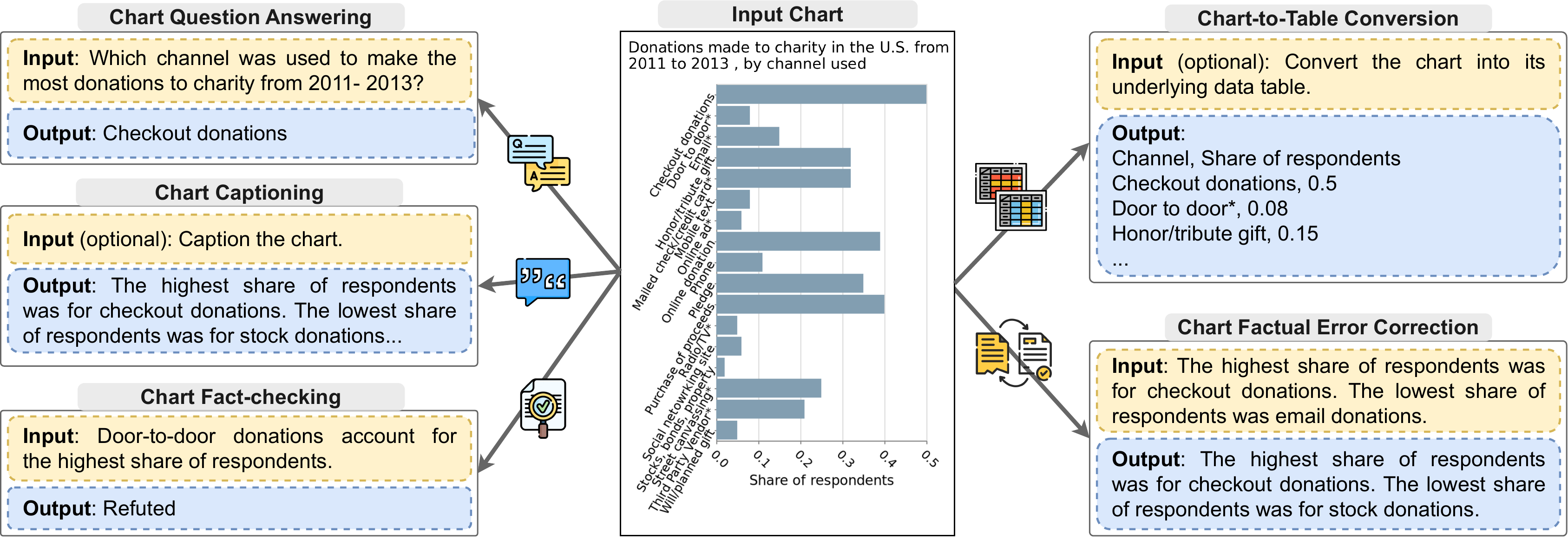}
 \vspace{-2mm}
 \caption{An overview of different chart understanding tasks. Each task is illustrated with one example.}
 \vspace{-5mm}
 \label{fig:chart_task_overview}
\end{figure*}

A wide variety of datasets have been developed for various tasks, including question answering, captioning, chart-to-table conversion, factual error correction, factual error correction, and fact-checking. An example of each task is displayed in \Cref{fig:chart_task_overview}.

\noindent \textbf{Chart Question Answering} involves presenting models with questions related to the content of a chart, which they must answer correctly. The challenge here is for the model to understand the trend of the underlying data and the relationship between data points. Two types of questions have been explored by prior work, factoid questions \cite{kafle2018dvqa, ebrahimikahou2018figureqa, methani2020plotqa, chaudhry2020leaf, singh-shekhar-2020-stl, masry-etal-2022-chartqa, chang2022mapqa, lu2023mathvista} and open-ended questions \cite{kantharaj-etal-2022-opencqa}. The answers to factoid questions are usually nouns (e.g. the values on the axes), verbs (e.g. increase or decrease), or adverbs (e.g. the magnitude of the trend), whereas the answers to open-ended questions are often of longer form, such as sentences.

\noindent \textbf{Chart Captioning}, also known as \textbf{Chart Summarization}, aims to generate a descriptive caption for a given visual representation \cite{obeid-hoque-2020-chart, kantharaj-etal-2022-chart, tang-etal-2023-vistext}. The generated caption should reflect the key insights or a summary of the information conveyed by the data visualization. 

\noindent \textbf{Chart-to-Table Conversion} requires a model to interpret the visual data representation and convert it into a tabular format \cite{methani2020plotqa, masry-etal-2022-chartqa, tang-etal-2023-vistext, liu-etal-2023-deplot}. This process involves extracting the data values and series from the chart and representing them in a structured table. 

\noindent \textbf{Chart Fact-Checking} involves verifying whether a given claim is factually consistent with the input chart, which helps identify cross-media misinformation \cite{fung-etal-2022-battlefront}. ChartFC \cite{akhtar-etal-2023-reading} and ChartCheck \cite{akhtar2023chartcheck} are the only chart fact-checking datasets. In contrast to the fact-checking literature, these two datasets only consider the \textit{support} or \textit{refute} labels and ignore the \textit{not enough information} label, where charts do not support or refute the corresponding claims. This setting is almost identical to the \textbf{Factual Inconsistency Detection for Chart Captioning} task \cite{huang2023lvlms}, in which the goal is to predict the relationship between the chart and the generated caption as \textit{consistent} (i.e. support) or \textit{inconsistent} (i.e. refute). The major difference is that chart fact-checking focuses on human-crated claims, whereas factual inconsistency detection uses machine-generated captions as text inputs.

\noindent \textbf{Chart Caption Factual Error Correction} is an extension of the fact-checking task where models are given a chart and a caption that may not be factually consistent with the chart. The goal is to identify and correct these factual errors, ensuring that the corrected caption faithfully represents the information presented in the charts. 

Due to the development of LVLMs, recent studies have introduced more challenging tasks. For example, tasks such as Chart Redrawing and Chart Referring QA have been proposed to push the boundaries of what these models can achieve \cite{xia-etal-2024-chartx, meng2024chartassisstant}. While these emerging tasks present exciting avenues for future research, they are still in their nascent stages and thus not comprehensively covered in this survey.

\subsection{Source}
\label{subsec:dataset_source}

In this subsection, we separately discuss the source of charts and the source of textual inputs (e.g. questions in chart question answering and claims in chart fact-checking).

\subsubsection{Source of Charts} Early datasets utilized visualization tools, such as Bokeh and Matplotlib, to create charts based on synthetic or real-world data. For example, DVQA \cite{kafle2018dvqa} and FigureQA \cite{ebrahimikahou2018figureqa} define a set of rules to generate the underlying data, while later efforts like PlotQA \cite{methani2020plotqa}, LEAF-QA \cite{chaudhry2020leaf}, and LEAF-QA++ \cite{singh-shekhar-2020-stl} began to source real-world data from various platforms, such as World Bank Open Data and World Development Indicators. More recently, more studies focused on collecting datasets with real-world charts. In particular, ChartQA \cite{masry-etal-2022-chartqa} and Chart-to-Text \cite{kantharaj-etal-2022-chart} gather charts from Statista and the Pew Research Center, while SciGraphQA \cite{li2023scigraphqa} and SciCap \cite{hsu-etal-2021-scicap-generating} collected charts from Arxiv. However, the latest developments, as shown in VisText \cite{tang-etal-2023-vistext}, revisit the use of visualization tools to accommodate the increasing demand for diverse visual layouts and scalability in data creation, especially for large instruction-tuning datasets (see \Cref{subsec:generation_based}). 

Additionally, with the advancement of LLMs, there is a growing trend of utilizing LLMs to produce synthetic charts. Examples include ChartLlama \cite{han2023chartllama}, ChartX \cite{xia-etal-2024-chartx}, and SimChart9K \cite{masry2024chartinstruct}, which simulate data across various topics and generate code to plot charts using LLMs.

The debate between the use of synthetic versus real-world charts is crucial in the chart understanding field. Synthetic charts, as generated by visualization tools according to predefined rules, offer the advantage of full control over the dataset, including the variety of chart types, the range and distribution of values, and the incorporation of specific challenges to test model robustness. This controlled setting allows for the systematic study of model behaviors and the identification of specific weaknesses. Furthermore, the process of generating synthetic charts is scalable and can easily produce large volumes of data necessary for training sophisticated machine learning models. However, the primary drawback of synthetic charts is their potential lack of realism. The generated charts may not fully capture the complexity and noise found in real-world charts, which can lead to poor performance in real-life scenarios for models trained on synthetic charts. Moreover, another challenge for automatically creating synthetic is balancing the desire for visual diversity and the need for the charts to remain comprehensible. We often aim to introduce variation in the styles of the charts, such as through randomized colors or plot types, to create a dataset that encompasses a wide range of visual appearances. However, ensuring that the resulting charts make visual sense poses a considerable difficulty. For instance, in a bar chart designed to compare different entities, if the bars corresponding to these entities are colored very similarly due to randomization, it may lead to confusion and impair the chart's readability. Such issues underscore the complexity of automatically generating synthetic charts that are not only diverse but also logically coherent and visually clear. \looseness=-1

Conversely, training on real-world charts could potentially lead to models better attuned to the nuances of real-world data visualizations. This could improve their performance on tasks outside controlled experimental settings. However, the use of real-world charts comes with its own set of challenges. %
First, collecting a large and diverse enough dataset to train robust models can be difficult and time-consuming. Additionally, the available charts may be biased towards certain domains, styles, or visual layouts, depending on the sources. Furthermore, labeling real-world charts for tasks such as question answering requires significant expert effort, making the dataset creation process slower and more expensive. \looseness=-1

Most datasets discussed in \Cref{tab:dataset_summary} are not in any particular domain. Charts from these sources (e.g. \textit{Wikimedia}, \textit{Pew Research Center}, \textit{Statista}, and etc) are about market research, public opinion, and demographic studies. Only a few datasets like SciCap \cite{hsu-etal-2021-scicap-generating} and PaperQA \cite{lu2023mathvista} dive into scientific domains. The lack of domain-specific chart understanding datasets not only underscores existing gaps in the field but also indicates great opportunities for future work. For example, domains such as healthcare, environmental science, and finance possess unique datasets with specialized chart types and data visualization needs. Further discussions regarding the lack of domain-specific chart understanding datasets can be found in \Cref{subsec:domain_specific_charts}.

\subsubsection{Source of Textual Inputs} Early efforts often employ template-based methods to generate textual inputs. FigureQA uses 15 templates to generate questions about charts without paraphrasing. The answers to these charts are binary (i.e. \textit{yes} or \textit{no}). DVQA increases the number of templates to 26 and allows answers to be either (1) derived from the 1,000 most frequent nouns in the Brown Corpus, %
or (2) extracted texts (e.g. label or value of a data point) from the chart. LEAF-QA generates question templates based on analytical reasoning questions in the Graduate Record Examination (GRE), resulting in a total of 35 question templates. LEAF-QA employs back-translation for automatic paraphrasing using the Google Translate API. LEAF-QA++ extends LEAF-QA by incrementing the number of question templates to 75 and including value-based questions where answers to these questions are the positions or values of data points within the charts. Concurrently, PlotQA includes open-vocabulary questions that require applying aggregation operations on the underlying chart data. 

While template-based methods for producing textual input queries allow for low-cost and efficient data generation processes, they are limited in textual richness and may not reflect real-world scenarios. Later work employs human-crafted textual inputs to better reflect real-world applications. ChartQA asks annotators to devise compositional (i.e. a combination of mathematical/logical operations) and visual (e.g. color and lengths) questions due to their high frequencies in real-world scenarios. Chart-to-Text curates human-written chart captions that are already publicly available on Statista\footnote{\url{https://www.statista.com/}} and Pew\footnote{\url{https://www.pewresearch.org/}}. %
VisText further enhances the semantic richness of the captions from Statista by collecting captions with more detailed descriptions regarding the statistical, perceptual, and cognitive characteristics of charts, such as trends and the relationships between chart elements. ChartCheck asks annotators to write claims that are supported or refuted by the chart while providing explanations.

More recently, due to the advancement of text generation, work like ChartX \cite{xia-etal-2024-chartx} and SimChart9K \cite{xia2023structchart} use LLMs to generate diverse and high-quality textual inputs.

\subsubsection{Plots}
In \Cref{tab:dataset_plot_types}, we provide a summary of the various types of plots that are included across datasets, illustrating the diversity of chart representations in visual data. The majority of the datasets commonly feature bar and line charts, which are fundamental in representing comparisons and trends across categories and time. For example, datasets such as FigureQA \cite{ebrahimikahou2018figureqa}, Chart2text \cite{obeid-hoque-2020-chart}, and Chocolate \cite{huang2023lvlms} all include these two chart types, underscoring their prevalence in the field. Pie charts are another ubiquitous plot type, offering a clear visual representation of proportional data. They are found in multiple datasets, such as LEAF-QA \cite{chaudhry2020leaf}, ChartQA \cite{masry-etal-2022-chartqa}, and ChartCheck \cite{akhtar2023chartcheck}, indicating that they are also a primary focus within the chart understanding fields.

While bar, line, and pie charts form the core of most datasets, some recent additions have begun to incorporate area plots, which are valuable for showing cumulative quantities and changes over time. Notably, OpenCQA \cite{kantharaj-etal-2022-opencqa} and Chart-to-Text \cite{kantharaj-etal-2022-chart} include area plots, thus expanding the scope of analysis possible with these datasets. Additionally, ChartX \cite{xia-etal-2024-chartx} is currently the most comprehensive chart understanding dataset, encompassing a total of 18 chart types ranging from bar charts to candlestick plots.

Specialized datasets often target less common but also important chart types. For example, MapQA \cite{chang2022mapqa} is dedicated to choropleth maps, showcasing the interest in geographical data representation. Moreover, datasets like PaperQA \cite{lu2023mathvista} go beyond conventional charts by including various forms of data visualization, such as t-SNE plots, potentially broadening the spectrum of chart understanding tasks.

\begin{table}[t]

    \small
    \centering
    \caption{Summary of the plot types available in each dataset.} 
    \begin{adjustbox}{max width=0.48\textwidth}
    {
    \begin{tabular}{lcccccc}
        \toprule
                         & \multicolumn{6}{c}{\textbf{Plot Types}}\\
                         \cmidrule(lr){2-7}
        \textbf{Dataset} & Line & Bar & Area & Pie & Scatter  & Others \\
        \midrule
        FigureQA \cite{ebrahimikahou2018figureqa} & \cmark~ & \cmark~ & \xmark~ &\cmark~ &\xmark~ & \xmark~\\
        DVQA \cite{kafle2018dvqa} & \xmark~ & \cmark~ &\xmark~ &\xmark~ &\xmark~ &\xmark~ \\
        LEAF-QA \cite{chaudhry2020leaf} & \cmark~ & \cmark~ & \xmark~ &\cmark~ &\xmark~ & \cmark~ \\
        LEAF-QA++ \cite{singh-shekhar-2020-stl} & \cmark~ & \cmark~ & \xmark~ &\cmark~ &\xmark~ & \cmark~ \\
        PlotQA \cite{methani2020plotqa} & \cmark~ & \cmark~ & \xmark~ &\xmark~ &\cmark~ & \xmark~ \\
        ChartQA \cite{masry-etal-2022-chartqa} & \cmark~ & \cmark~ & \xmark~ &\cmark~ &\xmark~ & \xmark~ \\
        MapQA \cite{chang2022mapqa} & \xmark~ & \xmark~ & \xmark~ &\xmark~ &\xmark~ & \cmark~ \\ %
        PaperQA \cite{lu2023mathvista} & \cmark~ & \cmark~ & \xmark~ &\cmark~ &\xmark~ & \cmark~ \\ %
        SciGraphQA \cite{li2023scigraphqa} & \cmark~ & \cmark~ & \cmark~ &\cmark~ &\cmark~ & \cmark~ \\
        MMC-Bench \cite{liu2023mmc} & \cmark~ & \cmark~ & \cmark~ &\cmark~ &\cmark~ & \cmark~ \\
        ChartBench \cite{xu2023chartbench} & \cmark~ & \cmark~ & \cmark~ &\cmark~ &\cmark~ & \cmark~ \\
        ArXivQA \cite{li2024multimodal} & \cmark~ & \cmark~ & \cmark~ &\cmark~ &\cmark~ & \cmark~ \\
        \midrule
        OpenCQA \cite{kantharaj-etal-2022-opencqa} & \cmark~ & \cmark~ & \cmark~ &\cmark~ &\cmark~ & \xmark~ \\ %
        \midrule
        FigCAP \cite{chen2019figure} & \cmark~ & \cmark~ & \xmark~ &\cmark~ &\xmark~ & \xmark~ \\
        Chart2text \cite{obeid-hoque-2020-chart} & \cmark~ & \cmark~ & \xmark~ &\xmark~ &\xmark~ & \xmark~ \\ %
        SciCap \cite{hsu-etal-2021-scicap-generating} & \cmark~ & \cmark~ & \cmark~ &\cmark~ &\cmark~ & \cmark~ \\
        LineCap \cite{mahinpei2022linecap} & \cmark~ & \xmark~ & \xmark~ &\xmark~ &\xmark~ & \xmark~\\
        ChaTa+ \cite{seweryn2021will} & \cmark~ & \cmark~ & \xmark~ &\cmark~ &\cmark~ & \cmark~ \\
        Chart-to-Text \cite{kantharaj-etal-2022-chart} & \cmark~ & \cmark~ & \cmark~ &\cmark~ &\cmark~ & \cmark~ \\ %
        ChartSumm \cite{rahman2023chartsumm} & \cmark~ & \cmark~ & \xmark~ &\cmark~ &\xmark~ & \xmark~ \\
        VisText \cite{tang-etal-2023-vistext} & \cmark~ & \cmark~ & \cmark~ &\xmark~ &\xmark~ & \xmark~ \\ %
        FigCaps-HF \cite{singh2023figcaps} & \cmark~ & \cmark~ & \cmark~ &\cmark~ &\cmark~ & \cmark~ \\
        ArXivCap \cite{li2024multimodal} & \cmark~ & \cmark~ & \cmark~ &\cmark~ &\cmark~ & \cmark~ \\
        \midrule
        ChartFC \cite{akhtar-etal-2023-reading} & \xmark~ & \cmark~ & \xmark~ &\xmark~ &\xmark~ & \xmark~  \\
        ChartCheck \cite{akhtar2023chartcheck} & \cmark~ & \cmark~ & \cmark~ &\cmark~ &\xmark~ & \xmark~ \\ %
        \midrule
        Chocolate \cite{huang2023lvlms} & \cmark~ & \cmark~ & \cmark~ &\cmark~ &\cmark~ & \cmark~ \\ %
        \midrule
        ChartLlama \cite{han2023chartllama} & \cmark~ & \cmark~ & \xmark~ &\cmark~ &\cmark~ & \cmark~\\
        ChartX \cite{xia-etal-2024-chartx} & \cmark~ & \cmark~ & \cmark~ &\cmark~ &\cmark~ & \cmark~ \\

        \bottomrule
    \end{tabular}
    }
    \end{adjustbox}
    
    \vspace{-5mm}
    \label{tab:dataset_plot_types}
    
\end{table}

\subsection{Evaluation Metric}
\label{subsec:metrics}
In this subsection, we illustrate the evaluation metrics for each task. A summary of the metrics for each task can be found in \Cref{tab:sota_performance}.%

\noindent \textbf{Factoid chart question answering.} The \textit{relaxed accuracy} metric proposed in \cite{methani2020plotqa} is usually used for this task. Relaxed accuracy (\texttt{RA}) checks if the output is identical to the ground truth, similar to an exact match. However, it allows $\varepsilon$\footnote{$\varepsilon$ is typically set to 5\%.} of numerical errors if the ground truth is a numerical value. %

{
\small
\begin{equation}
    \texttt{RA}(\hat{\mathbf{y}}, \mathbf{y}) = 
\begin{cases} 
  1 & \text{if } \hat{\mathbf{y}} = \mathbf{y}, \\
  1 & \text{if } \mathbf{y} \text{ is numerical } \land |\hat{\mathbf{y}} - \mathbf{y}| \leq \varepsilon \cdot |\mathbf{y}|, \\
  0 & \text{otherwise}
\end{cases}    
\end{equation}
}

Recently, due to the flexibility of large vision-language models (LVLMs) to take in diverse user instructions, studies propose \texttt{Acc+} to assess these models' chart understanding with contrasting prompts \cite{xu2023chartbench}. Specifically, for a given question $\mathbf{q}$ and the original ground truth answer $\mathbf{y}$, a \textit{wrong} answer $\mathbf{y'}$ is first randomly sampled from the metadata. %
Then, both answers are transformed into declarative statements on the question and integrated into the question. For example, with $\mathbf{q} =$ ``Which person has the highest share in company A?'', $\mathbf{y} =$ ``Alice'', and $\mathbf{y'} =$ ``Bob'', the prompts become $\mathbf{q^+} =$ ``Which person has the highest share in company A? \textit{Alice} has the highest share.'' and $\mathbf{q^-} =$ ``Which person has the highest share in company A? \textit{Bob} has the highest share.'', leading to binary labels $\mathbf{y^+} =$ ``yes'' and $\mathbf{y^-} =$ ``no''. The model is then assessed on its ability to distinguish between these contrasting prompts' consistency with the chart, with \texttt{Acc+} calculated as:

{
\small
\begin{equation}
    \texttt{Acc+}(\mathbf{y^+}, \mathbf{y^-}, \hat{\mathbf{y^+}}, \hat{\mathbf{y^-}}) = 
\begin{cases} 
  1 & \text{if } \hat{\mathbf{y^+}} = \mathbf{y^+} \text{and } \hat{\mathbf{y^-}} = \mathbf{y^-}, \\
  0 & \text{otherwise}
\end{cases}    
\end{equation}
}
Note that this setting resembles factual inconsistency detection \cite{huang2023lvlms} and fact-checking \cite{akhtar2023chartcheck}. However, it is distinct in its use of contrasting prompts and the framing of textual queries as question-answer pairs instead of declarative statements.

\noindent \textbf{Chart-to-Table Conversion.} Three metrics have been developed for the chart-to-table conversion task, %
\textit{Relative Number Set Similarity} (\texttt{RNSS}) \cite{masry-etal-2022-chartqa}, \textit{Relative Mapping Similarity} (\texttt{RMS}) \cite{liu-etal-2023-deplot}, and \textit{Structuring Chart-oriented Representation Metric} (SCRM) \cite{xia2023structchart}. For this task, we can view tables as unordered collections of mappings from a row-and-column header $(r, c)$ to a single value $v$. \texttt{RNSS} represents each entry of the predicted table using the values only. Specifically, $\{ \hat{\mathbf{T}}^v_i\}_{1 \leq i \leq N}$ and $\{ \mathbf{T}^v_j\}_{1 \leq j \leq M}$ represent the set of values in the ground truth and generated tables, respectively. \texttt{RNSS} defines the similarity $S$ between each pair of values:

{
\small
\begin{align}
    S(\hat{\mathbf{T}}_i, \mathbf{T}_j) = 1 - \min \left( 1,  \frac{| \hat{\mathbf{T}}^{v}_i - \mathbf{T}^{v}_j |}{|\mathbf{T}_j|}\right).
\end{align}
}

Then, a minimal cost matching $\mathbf{X}^{\texttt{RNSS}} \in \{0, 1\}^{N \times M}$ between $\{ \hat{\mathbf{T}}_i\}_{1 \leq i \leq N}$ and $\{ \mathbf{T}_j\}_{1 \leq j \leq M}$ is obtained based on $S$ and a cost optimization matching algorithm $\mathcal{M}$:

{
\small
\begin{align}
C^{\texttt{RNSS}} &= 1 - S(\hat{\mathbf{T}}, \mathbf{T}), \\
\mathbf{X}^{\texttt{RNSS}} &= \mathcal{M}(C^{\texttt{RNSS}}).
\end{align}
}
Finally, the \texttt{RNSS} score is computed as 

{
\small
\begin{align}
    \texttt{RNSS} = 1 - \frac{\sum^{N}_{i=1} \sum^{M}_{j=1} \mathbf{X}^{\texttt{RNSS}}_{i,j} (1 - S(\hat{\mathbf{T}}_i, \mathbf{T}_j)) }{\max (N, M )}.
\end{align}
}

However, the \texttt{RNSS} metric exhibits a number of shortcomings. First, it is unable to discern the position of values across the table, thus failing to capture the layout information. Second, it disregards any non-numeric content within the table, which may hold critical contextual relevance. Third, \texttt{RNSS} disproportionately rewards estimates with substantial relative errors, potentially undermining the robustness of the evaluation. Lastly, it does not consider precision and recall.

To address these limitations, \texttt{RMS} represents each entry of the predicted table as $\hat{\mathbf{T}}_i = ( \hat{\mathbf{T}}^{r,c}_i, \hat{\mathbf{T}}^v_i )$ and that of the ground truth table as $\mathbf{T}_j = ( \mathbf{T}^{r,c}_j, \mathbf{T}^v_j )$. 
Here, $\hat{\mathbf{T}}^{r,c}_i$ denotes the string concatenation of row header $\hat{\mathbf{T}}^{r}_i$ and column header $\hat{\mathbf{T}}^{c}_i$. The distance between two keys is computed using Normalized Levenshtein Distance \cite{levenshtein1966binary}, denoted as $\texttt{NL}$.

{
\small
\begin{equation}
    S^{\text{key}}_{\tau}(\hat{\mathbf{T}}_i, \mathbf{T}_j) = 
    \begin{cases}
        0 & \text{if } \texttt{NL}(\hat{\mathbf{T}}^{r,c}_i, \mathbf{T}^{r,c}_j ) > \tau, \\
        1 - \texttt{NL}(\hat{\mathbf{T}}^{r,c}_i , \mathbf{T}^{r,c}_j ) & \text{otherwise}.\\
    \end{cases}
\end{equation}
}
In scenarios where the Normalized Levenshtein Distance exceeds the threshold $\tau$, the similarity score $S^{\text{key}}_{\tau}$ is set to 0 to avoid allocating partial credits to generated tables that are significantly dissimilar to ground truth tables. Similarly, a separate parameter $\theta$ is introduced to prevent dissimilar values getting partial credits. The similarity between two values $S^{\text{value}}_{\theta}$ is defined as:

{
\small
 \begin{equation}   
S^{\text{value}}_{\theta}(\hat{\mathbf{T}}_i, \mathbf{T}_j) = 
\begin{cases} 
0 & \text{if } \frac{|\hat{\mathbf{T}}^v_i - \mathbf{T}^v_j|}{|\mathbf{T}^v_j|} > \theta, \\
1 - \min\left( 1,  \frac{|\hat{\mathbf{T}}^v_i - \mathbf{T}^v_j|}{|\mathbf{T}^v_j|}\right) & \text{otherwise}.
\end{cases}
 \end{equation}
}

The similarity between two mappings $S_{\tau, \theta}(\hat{\mathbf{T}}_i, \mathbf{T}_j)$ is the product of the key similarity and value similarity:

{
\small
\begin{align}
    S_{\tau, \theta}(\hat{\mathbf{T}}_i, \mathbf{T}_j) &= S^{\text{key}}_{\tau}(\hat{\mathbf{T}}_i, \mathbf{T}_j) \cdot S^{\text{value}}_{\theta}(\hat{\mathbf{T}}_i, \mathbf{T}_j), 
\end{align}
}
where $\tau$ and $\theta$ are set to 0.5 and 0.1, respectively. Then, a minimal cost matching $\mathbf{X}_{\texttt{RMS}} \in \{0, 1\}^{N \times M}$ is found using the cost function $C^{\texttt{RMS}}$,

{
\small
\begin{align}
C^{\texttt{RMS}} &= 1 - S_{\tau, \theta}(\hat{\mathbf{T}}, \mathbf{T}), \\
\mathbf{X}^{\texttt{RMS}} &= \mathcal{M}(C^{\texttt{RMS}}).
\end{align}
}
With $\mathbf{X}^{\texttt{RMS}}$ calculated, precision and recall can be computed as follows:

{
\small
\begin{align}
\texttt{RMS}_{\text{precision}} &= \frac{\sum^{N}_{i=1}\sum^{M}_{j=1} \mathbf{X}^{\texttt{RMS}}_{ij}S_{\tau, \theta}(\hat{\mathbf{T}}_i, \mathbf{T}_j)}{N},\\
\texttt{RMS}_{\text{recall}} &= \frac{\sum^{N}_{i=1}\sum^{M}_{j=1} \mathbf{X}^{\texttt{RMS}}_{ij}S_{\tau, \theta}(\hat{\mathbf{T}}_i, \mathbf{T}_j)}{M}.
\end{align}
}
The F1 score is the harmonic mean of $\texttt{RMS}_{\text{precision}}$ and $\texttt{RMS}_{\text{recall}}$. \footnote{Note that the equations for \texttt{RMS} are different from the original paper \cite{liu-etal-2023-deplot}. This is because there is an inconsistency between the equations listed in the paper and the corresponding code implementation. We rewrite all the equations by following their source code.}

Unlike \texttt{RNSS} and \texttt{RMS}, \texttt{SCRM} represents each table entry as a triplet through the Structured Triplet Representations (STR) framework. STR is designed to efficiently and robustly represent the complex data relations within chart information by capturing the relationships between row and column headers and their corresponding values. SCRM is computed by first calculating entity and value matching metrics. Thus, the SCRM metric provides a comprehensive evaluation metric to assess the performance of chart perception from structured information extraction, addressing the inherent complexities associated with data relations in visual charts.

\noindent \textbf{Chart Fact-checking and Factual Inconsistency Detection for Chart Captioning.} %
Macro F1 and accuracy are used for chart fact-checking, while Kendall's Tau \cite{kendall1938new} is adopted for the factual inconsistency detection for chart captioning task.

\noindent \textbf{Long-form Chart Understanding Tasks.} The tasks %
discussed above are relatively easy to evaluate since the ground truths are either short answers or long answers that can be easily decomposed into short answers due to the structural nature of the outputs. Evaluating tasks that involve long-form texts, including \textbf{Open-ended Chart Question Answering}, \textbf{Chart Captioning}, and \textbf{Chart Caption Factual Error Correction} %
is more challenging since the outputs are much longer and cannot be easily broken down into short sub-outputs. Most of the prior work use reference-based evaluation metrics, which score the outputs against reference captions or answers. Some adopt lexical-based metrics, such as ROUGE \cite{lin-2004-rouge}, BLEU \cite{papineni-etal-2002-bleu}, and CIDEr \cite{Vedantam_2015_CIDEr}, another line of studies use semantic-focused metrics like perplexity and BERTScore \cite{zhang-etal-2020-bertscore}, and BLEURT \cite{sellam-etal-2020-bleurt}, others use Large Language Models (LLMs) as evaluation metrics \cite{han2023chartllama}, motivated by the text generation fields \cite{liu-etal-2023-g, huang2023embrace}. %
However, these metrics have several drawbacks. First, these metrics do not correlate well with \textit{faithfulness}, or \textit{factual consistency}, %
between the chart and the outputs, as demonstrated by prior work in the field of summarization \cite{wang-etal-2020-asking, huang-etal-2023-swing, qiu2023amrfact}. %
Second, they depend on the quality (e.g. \textit{coverage}) %
of the reference outputs. Taking chart captioning as an example, the majority of instances in existing datasets do not cover all key insights from the chart. Therefore, it is possible that the generated caption is faithful to the given charts but has a low lexical overlap with the reference caption. To address these issues, \textsc{ChartVE} \cite{huang2023lvlms} formulates the factual inconsistency detection problem as a visual entailment task. \textsc{ChartVE} learns to predict the entailment probability $E(\mathbf{c}, \mathbf{q})$ from the input chart $\mathbf{c}$ to a caption sentence $\mathbf{q}$. The \textsc{ChartVE} score is computed by taking the minimum of all caption sentences $\mathbf{q} \in \mathbf{Q}$:

{
\small
\begin{align}
\textsc{ChartVE}= \min_{\mathbf{q} \in \mathbf{Q}} E(\mathbf{c}, \mathbf{q}).
\end{align}
}
In detecting factual inconsistency between charts and captions, \textsc{ChartVE} demonstrates better performance than open-source Large Vision-language Models (LVLMs), such as LLaVA-V1.5 \cite{liu2023improvedllava} and compares favorably to proprietary LVLMs like GPT-4V \cite{openai2023gpt4v}. However, apart from \textit{factual inconsistency}, there is a lack of research on the development of metrics for evaluating other criteria. In \Cref{subsec:eval}, we discuss the importance of other aspects, such as \textit{coverage} and \textit{relevancy}.

\section{Modeling Strategies} \label{Sec:4}

\definecolor{connect-line}{RGB}{0,0,0}
\definecolor{middle-color}{RGB}{255,255,255}
\definecolor{leaf-color}{RGB}{255,255,255}
\definecolor{line-color}{RGB}{25,25,112}

\definecolor{black}{RGB}{0,0,0}

\definecolor{pure}{RGB}{112,25,25}
\definecolor{node}{RGB}{25,25,112}
\definecolor{graph}{RGB}{25,112,25}

\tikzstyle{pure-leaf}=[draw=pure,
    rounded corners,minimum height=1em,
    fill=leaf-color!40,text opacity=1, align=center,
    fill opacity=.5,  text=black,align=left,font=\scriptsize,
    inner xsep=3pt,
    inner ysep=1pt,
]
\tikzstyle{pure-middle}=[draw=pure,
    rounded corners,minimum height=1em,
    fill=middle-color!40,text opacity=1, align=center,
    fill opacity=.5,  text=black,align=left,font=\scriptsize,
    inner xsep=3pt,
    inner ysep=1pt,
]
    
\tikzstyle{node-leaf}=[draw=node,
    rounded corners,minimum height=1em,
    fill=leaf-color!40,text opacity=1, align=center,
    fill opacity=.5,  text=black,align=left,font=\scriptsize,
    inner xsep=3pt,
    inner ysep=1pt,
]
\tikzstyle{node-middle}=[draw=node,
    rounded corners,minimum height=1em,
    fill=middle-color!40,text opacity=1, align=center,
    fill opacity=.5,  text=black,align=left,font=\scriptsize,
    inner xsep=3pt,
    inner ysep=1pt,
]

\tikzstyle{graph-leaf}=[draw=graph,
    rounded corners,minimum height=1em,
    fill=leaf-color!40,text opacity=1, align=center,
    fill opacity=.5,  text=black,align=left,font=\scriptsize,
    inner xsep=3pt,
    inner ysep=1pt,
]
\tikzstyle{graph-middle}=[draw=graph,
    rounded corners,minimum height=1em,
    fill=middle-color!40,text opacity=1, align=center,
    fill opacity=.5,  text=black,align=left,font=\scriptsize,
    inner xsep=3pt,
    inner ysep=1pt,
]

\tikzstyle{leaf}=[draw=line-color,
    rounded corners,minimum height=1em,
    fill=leaf-color!40,text opacity=1, align=center,
    fill opacity=.5,  text=black,align=left,font=\scriptsize,
    inner xsep=3pt,
    inner ysep=1pt,
    ]
\tikzstyle{middle}=[draw=line-color,
    rounded corners,minimum height=1em,
    fill=middle-color!40,text opacity=1, align=center,
    fill opacity=.5,  text=black,align=left,font=\scriptsize,
    inner xsep=3pt,
    inner ysep=1pt,
    ]
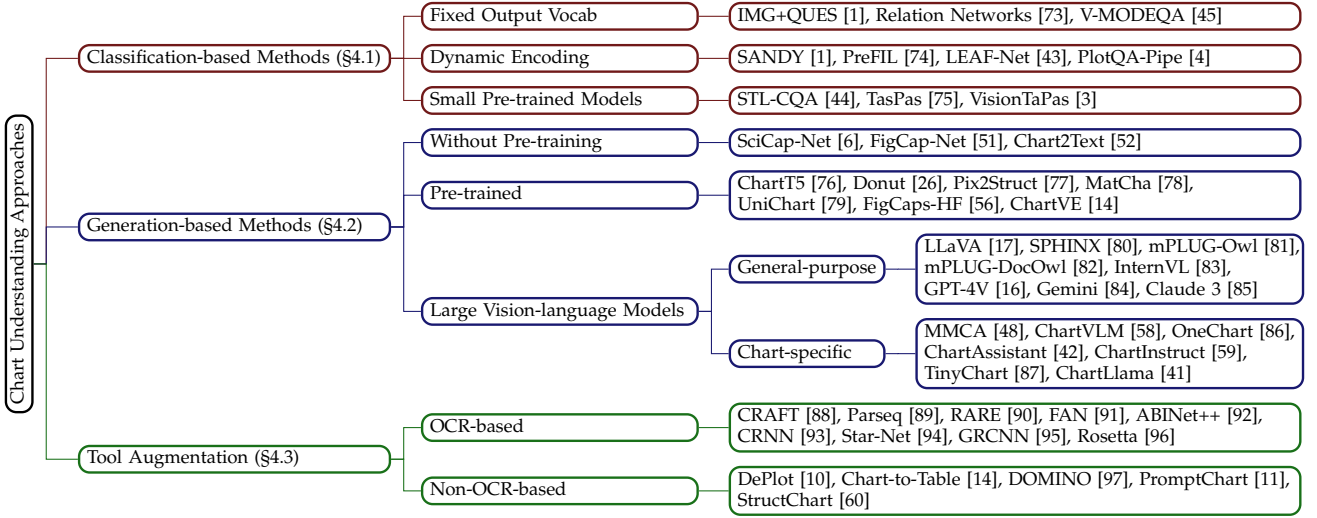
\begin{figure*}[ht]
\centering
\begin{forest}
  for tree={
    forked edges,
    grow=east,
    reversed=true,
    anchor=base west,
    parent anchor=east,
    child anchor=west,
    base=middle,
    font=\scriptsize,
    rectangle,
    line width=0.9pt,
    draw=connect-line,
    rounded corners,align=left,
    minimum width=2em,
    s sep=5pt,
    inner xsep=3pt,
    inner ysep=1pt,
  },
  where level=1{text width=4.5em}{},
  where level=2{text width=6em,font=\scriptsize}{},
  where level=3{font=\scriptsize}{},
  where level=4{font=\scriptsize}{},
  where level=5{font=\scriptsize}{},
  [
                Chart Understanding Approaches, black,rotate=90,anchor=north,edge=pure
                [
                   Classification-based Methods (\S \ref{subsec:classification_based}) , pure-middle, edge=pure, text width=11.7em
                   [
                        Fixed Output Vocab, pure-middle, edge=pure, text width=10.3em
                        [
                            IMG+QUES \cite{kafle2018dvqa}{,}
                            Relation Networks \cite{santoro2017simple}{,}
                            V-MODEQA \cite{chang2022mapqa}
                            , pure-leaf, text width=22em, edge=pure
                        ]
                   ]
                   [
                        Dynamic Encoding, pure-middle, edge=pure, text width=10.3em
                        [
                            SANDY \cite{kafle2018dvqa}{,}
                            PreFIL \cite{kafle2020answering}{,}
                            LEAF-Net \cite{chaudhry2020leaf}{,}
                            PlotQA-Pipe \cite{methani2020plotqa}
                            , pure-leaf, text width=22em, edge=pure
                        ]
                   ]
                   [
                        Small Pre-trained Models, pure-middle, edge=pure, text width=10.3em
                        [
                            STL-CQA \cite{singh-shekhar-2020-stl}{,}
                            TasPas \cite{herzig-etal-2020-tapas}{,}
                            VisionTaPas \cite{masry-etal-2022-chartqa}
                            , pure-leaf, text width=22em, edge=pure
                        ]
                   ]
                ]
                [
                    Generation-based Methods (\S \ref{subsec:generation_based}), node-middle, edge=node, text width=11.7em
                    [
                        Without Pre-training, node-middle, edge=node, text width=10.3em
                        [
                            SciCap-Net \cite{hsu-etal-2021-scicap-generating}{,}
                            FigCap-Net \cite{chen2019figure}{,}
                            Chart2Text \cite{obeid-hoque-2020-chart}
                            , node-leaf, text width=22em, edge=node
                        ]
                    ]
                    [    
                        Pre-trained, node-middle, edge=node, text width=10.3em
                        [
                            ChartT5 \cite{zhou-etal-2023-enhanced}{,}
                            Donut \cite{kim2022donut}{,}
                            Pix2Struct \cite{lee2023pix2struct}{,}
                            MatCha \cite{liu-etal-2023-matcha}{,}\\
                            UniChart \cite{masry-etal-2023-unichart}{,}
                            FigCaps-HF \cite{singh2023figcaps}{,}
                            ChartVE~\cite{huang2023lvlms}
                            , node-leaf, text width=22em, edge=node
                        ]
                    ]
                    [    
                        Large Vision-language Models, node-middle, edge=node, text width=10.3em
                        [
                            General-purpose, node-middle, edge=node, text width=5.5em
                            [
                                LLaVA~\cite{liu2023improvedllava}{,}
                                SPHINX~\cite{lin2023sphinx}{,}
                                mPLUG-Owl~\cite{ye2023mplug}{,}\\
                                mPLUG-DocOwl \cite{ye2023mplugdocowl}{,}
                                InternVL \cite{chen2024far}{,}\\
                                GPT-4V \cite{openai2023gpt4v}{,}
                                Gemini \cite{team2023gemini}{,}
                                Claude 3 \cite{anthropic2024claude3}
                                , node-leaf, text width=14.7em, edge=node
                            ]
                        ]
                        [
                            Chart-specific, node-middle, edge=node, text width=5.5em
                            [
                                MMCA~\cite{liu2023mmc}{,}
                                ChartVLM \cite{xia-etal-2024-chartx}{,}
                                OneChart \cite{chen2024onechart}{,}\\
                                ChartAssistant~\cite{meng2024chartassisstant}{,}
                                ChartInstruct~\cite{masry2024chartinstruct}{,}\\
                                TinyChart \cite{zhang2024tinychart}{,}
                                ChartLlama~\cite{han2023chartllama}
                                , node-leaf, text width=14.7em, edge=node
                            ]
                        ]
                    ]
                ]
                [
                    Tool Augmentation (\S \ref{subsec:tool_aug}), graph-middle, edge=graph, text width=11.7em
                    [
                        OCR-based, graph-middle, edge=graph, text width=10.3em
                        [
                            CRAFT \cite{baek2019character}{,}
                            Parseq \cite{bautista2022scene}{,}
                            RARE \cite{shi2016robust}{,}
                            FAN \cite{cheng2017focusing}{,}
                            ABINet++ \cite{fang2022abinet++}{,}\\
                            CRNN \cite{shi2016end}{,}
                            Star-Net \cite{liu2016star}{,}
                            GRCNN \cite{wang2017gated}{,}
                            Rosetta \cite{borisyuk2018rosetta}
                            ,graph-leaf, text width= 22em, edge=graph
                        ]
                    ]
                    [
                        Non-OCR-based, graph-middle, edge=graph, text width=10.3em
                        [
                            DePlot~\cite{liu-etal-2023-deplot}{,}
                            Chart-to-Table~\cite{huang2023lvlms}{,}
                            DOMINO~\cite{wang2023domino}{,}
                            PromptChart~\cite{do2023llms}{,}\\
                            StructChart~\cite{xia2023structchart}
                            ,graph-leaf, text width= 22em, edge=graph
                        ]
                    ]
                ]
    ]  
\end{forest}
\vspace{-2mm}
\caption{A taxonomy of chart understanding approaches with representative work.}
\label{dia:model_taxonomy}
\vspace{-5mm}
\end{figure*}

\Cref{dia:model_taxonomy} shows an overview of different chart understanding approaches. In this section, we delve into the modeling strategies that have been influential in the development of chart understanding. This exploration spans classification-based and generation-based models, distinguishing between approaches that rely on pre-training and those that do not. Additionally, we highlight foundation models, particularly Large Vision-Language Models (LVLMs), as a pivotal development in the field, showcasing their capacity to ingest and interpret complex visual data through advanced encoding techniques. Finally, the section explores the concept of tool augmentation, illustrating how external systems can enhance model performance by decomposing the perception and reasoning capabilities of chart understanding tasks.

\subsection{Classification-based Models}
\label{subsec:classification_based}
Classification-based methods have mainly tackle the task of chart question answering, leveraging a combination of visual and textual features to understand and interpret different types of charts. This subsection provides an overview of the evolution and key strategies employed in classification-based chart understanding.
\subsubsection{Fixed Output Vocabulary and Initial Approaches} Early methods, such as IMG+QUES \cite{kafle2018dvqa}, utilized a fixed, limited vocabulary in the output layer. This approach typically involved the use of LSTM networks for encoding questions and a CNN architecture, like ResNet, for encoding chart figures. The representations obtained from both modalities were then fused, generally through concatenation of feature vectors, and passed through a multi-layer perceptron to predict the answer, often relying on a softmax output layer for the final answer classification. Concretely,

\begin{equation}
    \hat{\mathbf{y}} = \argmax_i \left( \text{softmax}(\mathbf{W} \cdot \mathbf{H}_{\text{concat}} + \mathbf{b}) \right),
\end{equation}
where $\mathbf{H}_{\text{concat}} = \text{CONCAT}(\mathbf{H}_{c}, \mathbf{H}_{q})$ signifies the concatenated feature vector derived from the vision encoder $\mathbf{H}_{c}$ and the question representation $\mathbf{H}_{q}$, $\mathbf{W}$ and $\mathbf{b}$ are the weights and bias of the output layer. The vocabulary for the output layer is usually constructed by gathering unique tokens from the answers in the training set. This leads to a significant limitation of these early models: the out-of-vocabulary (OOV) challenge, where the answers in the test set contain tokens that do not overlap with those in the training set. DVQA is one of the early datasets that explores this issue. 

\subsubsection{Dynamic Encoding and Addressing OOV Challenges} To overcome the limitations associated with fixed vocabularies, subsequent work introduced dynamic encoding schemes. SANDY\cite{kafle2018dvqa}, for instance, enhanced chart QA capabilities by integrating a Stacked Attention Network (SAN) \cite{yang2016stacked} with an OCR-based sub-network. This dual-sub-network approach allowed SANDY to generate generic answers and, simultaneously, identify and interpret textual and numerical chart elements like axis labels and data values. By constructing an image-specific dictionary linking spatial positions of chart elements to dictionary entries, SANDY successfully tackled the OOV challenge, allowing for the dynamic generation of chart-specific answers based on the content identified within the chart.

Similarly, PReFIL \cite{kafle2020answering} advances the understanding of charts by jointly learning bimodal embeddings from both low-level and high-level image features. This comprehensive understanding aids in answering complex questions that require advanced reasoning. Concurrently, LEAF-Net \cite{chaudhry2020leaf} improves chart understanding ability by employing a Mask R-CNN network \cite{he2017mask} for element identification and an attention mechanism to focus on relevant chart areas based on the encoded question, thus enhancing answer prediction accuracy.

\subsubsection{Enhancements Through Pre-training}
Inspired by the success of text pre-training models, such as BERT \cite{devlin-etal-2019-bert} and BART \cite{lewis-etal-2020-bart}, more recent work introduced pre-training to further improve the performance of classification-based chart understanding methods like STL-CQA \cite{singh-shekhar-2020-stl}. These approaches utilize several pre-training objectives tailored for enhancing model understanding of chart images and associated texts. Pre-training enables these models to learn more robust and comprehensive feature representations, improving their ability to interpret charts accurately and respond to more complex queries effectively. %

\subsubsection{Assumptions on OCR Availability} It is essential to note that irrespective of the advancements in classification-based methods, there remains an underlying assumption that predicted or ground-truth OCR information is available. This assumption plays a critical role in enabling these models to identify and understand textual elements within charts, which is crucial for accurately answering questions about the charts. The reliance on external OCR systems is eliminated due to the rise of OCR-free, end-to-end document understanding models like Donut \cite{kim2022donut} and Pix2Struct \cite{lee2023pix2struct}, which will also be discussed in \Cref{subsec:generation_based}.

\begin{figure*}[t]
 \centering
 \includegraphics[width=\linewidth]{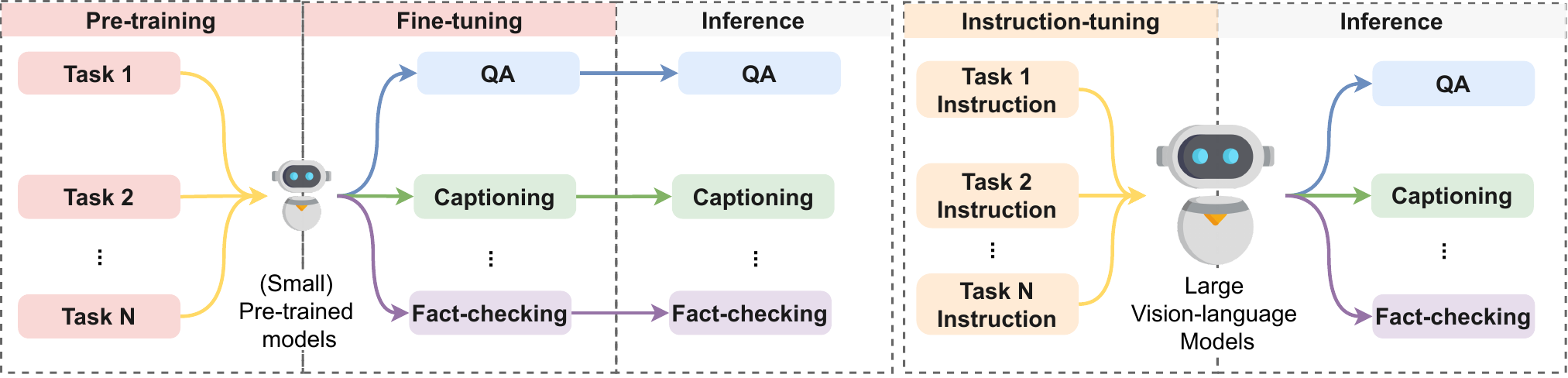}
 \vspace{-5mm}
 \caption{A comparison between (small) pre-trained vision-language models and LVLMs. In addition to the scale of models, the biggest difference between these two types of models is that LVLMs do not need task-specific fine-tuning since instruction-tuning allows them to generalize to unseen tasks.}
 \vspace{-5mm}
 \label{fig:pretrained_vs_lvlms}
\end{figure*}

\subsection{Generation-based Models}
\label{subsec:generation_based}

Classification-based approaches are limited to tasks with fixed vocabulary or single-token outputs, unsuitable for many chart understanding tasks such as those involving long-form text outputs. Hence, all recent chart understanding methods adopt generation-based architectures.%
These methods move toward more integrated, end-to-end trainable frameworks. This subsection delineates the evolution and distinctive characteristics of these models, focusing on their encoder and decoder architecture and pre-training objectives.

\subsubsection{Non-pretrained Approaches} Initially, generation-based models combined classical image and text processing techniques to understand charts and generate descriptions. They relied on separate modules to handle different modalities without any form of pre-training, using a CNN for visual encoding and an RNN for sequential text generation. Models like the combination of a ResNet \cite{he2016deep} or DenseNet \cite{huang2017densely} chart encoder with an LSTM \cite{hochreiter1997long} decoder \cite{hsu-etal-2021-scicap-generating, chen2019figure, mahinpei2022linecap} represent these early endeavors. Image features extracted by the CNN were fed into the RNN to generate text descriptions auto-regressively. While effective, the lack of pre-trained knowledge and modality integration limited their performance and generalizability.

Data-to-text models like Chart2text \cite{obeid-hoque-2020-chart} and the Field-Infusing Model \cite{chen-etal-2020-logical} represented a step toward more data-focused approaches, assuming access to underlying chart data or requiring external OCR for data extraction. These approaches aimed at directly converting structured data into textual summaries or captions, using Transformer-based architectures to better capture data relations and semantics. However, these methods still operated without the benefits of pre-training, limiting their ability to leverage large-scale data for improved understanding and generation.

\subsubsection{Small Pre-trained Models} The advent of pre-trained models significantly enhanced the chart understanding capabilities of VL models. ChartT5 \cite{zhou-etal-2023-enhanced} emerged as a pioneering approach that utilizes pre-training objectives to enhance its understanding of charts and tables. By recovering masked information in a table using chart images, ChartT5 addressed the extraction and reasoning tasks more effectively. However, it still assumed the presence of a ground truth data table or relied on external OCR systems, highlighting a dependency that later models sought to eliminate.

Significantly advancing the field, end-to-end visual document understanding models like Donut \cite{kim2022donut} and Pix2Struct \cite{lee2023pix2struct} introduced the capability to directly parse textual information from images, removing the need for separate OCR processes. Donut utilizes a Swin Transformer \cite{liu2021swin} for the image encoding and the decoder of mBART \cite{liu-etal-2020-mbart} for decoding, focusing on natural image parsing to produce text sequences from images. Pix2Struct, employing a Vision Transformer (ViT) \cite{dosovitskiy2020image} as the image encoder and a Transformer-based decoder. The pre-training objective is screenshot parsing, which learns to predict an HTML-based parse from a screenshot of a webpage, providing a structured training signal that encapsulates both textual and layout information. There are two major distinctions between Donut and Pix2Struct: (1) screenshot parsing may allow the model to learn the structure of the image better since HTML provides stronger signals about the layouts of the input image. (2) Donut is multilingual since it uses mBART's decoder for decoding texts.

More recent studies further enhanced chart understanding performance by continuing pre-training these end-to-end visual document understanding models given their advanced capabilities in the layout of images. For instance, MatCha \cite{liu-etal-2023-matcha} builds upon Pix2Struct and undergoes pre-training with three chart-specific objectives: chart-to-table conversion, chart-to-code conversion, and mathematical question answering. These objectives not only enhance the model's ability to interpret values and the structure of the chart but also enhance its proficiency in generating precise, contextually relevant content. Furthermore, to prevent catastrophic forgetting of the screenshot parsing capability, MatCha includes the screenshot parsing task, the original pre-training objective of Pix2Struct, as its fourth pre-training objective. Similarly, UniChart \cite{masry-etal-2023-unichart} is learned by continuing pre-training Donut with four pre-training objectives, including chart-to-text conversion, chart captioning, numerical and visual reasoning, and open-ended question answering. This broad spectrum of pre-training objectives significantly broadens UniChart’s capabilities, enabling it to tackle a wide array of chart-related tasks with improved overall text generation quality \cite{huang2023lvlms}. With the advancement in chart understanding achieved by these pre-trained models, we have the following discussions.

First, \textbf{do we still need external OCR systems?} The advancements in chart understanding and interpretation achieved by UniChart and MatCha over ChartT5 highlight a pivotal shift in the field towards leveraging OCR-free models such as Pix2Struct and Donut. These developments underscore a crucial insight: the integration of advanced OCR-free pre-training objectives can obviate the dependency on traditional OCR systems, which often represent a bottleneck in terms of accuracy and adaptability. UniChart and MatCha exhibit superior performance without needing the direct input of ground truth tables, which ChartT5 relies upon. This clearly shows the increased efficacy and robustness of models developed using the OCR-free paradigm. %

Second, \textbf{is it important \textit{not} to forget knowledge learned during the previous pre-training stage?} The distinct approaches adopted by UniChart and MatCha regarding their pre-training objectives throw light on nuanced strategies for improving chart understanding capabilities. UniChart forgoes the incorporation of Donut's original pre-training objective, only focusing instead on a diverse set of tailored chart-centric tasks. Conversely, MatCha includes Pix2Struct's original pre-training objective to address the potential issue of catastrophic forgetting and ensures a broader knowledge base for the model. The fact that UniChart exhibits much better performance despite excluding the pre-training objectives used by Donut \cite{masry-etal-2023-unichart, huang2023lvlms} is indicative of the importance of designing diverse pre-training objectives to closely align with the downstream chart understanding tasks. More importantly, this suggests that forgetting knowledge learned during the previous pre-training stage may not influence downstream performance significantly.

\begin{table*}[t]
    \small
    \centering
    \caption{Summary of various pre-trained and instruction fine-tuned vision-language models for chart understanding. Note that MMCA is instruction-tuned from mPLUG-Owl, ChartLlama is instruction-tuned from LLaVA-1.5, and ChartAssistant-S is instruction-tuned from SPHINX. The weights of ChartInstruct's vision encoder are initialized from UniChart. }
    \begin{adjustbox}{max width=0.95\textwidth}
    {
    \begin{tabular}{lcccc}
        \toprule
        \textbf{Model} & \textbf{Chart Encoder} & \textbf{Text Encoder} & \textbf{Text Decoder} & \textbf{\# Parameters} \\
        \midrule
        \multicolumn{5}{c}{\textit{(Small) Pre-trained Models}} \\
        \midrule
        ChartT5 \cite{zhou-etal-2023-enhanced} & Mask R-CNN \cite{he2017mask} & T5's encoder \cite{raffel2020exploring} & T5's decoder \cite{raffel2020exploring} & 224M \\
        Pix2Struct \cite{lee2023pix2struct} & $\text{ViT-B}_{\text{16}}$/$\text{ViT-L}_{\text{16}}$ \cite{dosovitskiy2020image} & - & Transformer \cite{DBLP:conf/nips/Transformer17} & 282M/1.3B \\
        MatCha \cite{liu-etal-2023-matcha} & $\text{ViT-B}_{\text{16}}$ \cite{dosovitskiy2020image} & - & Transformer \cite{DBLP:conf/nips/Transformer17} &  282M \\
        Donut \cite{kim2022donut} & Swin Transformer \cite{liu2021swin} & - & mBART's decoder \cite{liu-etal-2020-mbart} & 201M \\
        UniChart \cite{masry-etal-2023-unichart} & Swin Transformer \cite{liu2021swin}  &  - & mBART's decoder \cite{liu-etal-2020-mbart} & 201M \\
        ChartAssistant-D \cite{meng2024chartassisstant} & Swin Transformer \cite{liu2021swin} & - & mBART's decoder \cite{liu-etal-2020-mbart} & 201M \\
        ChartInstruct-Flan-T5-XL \cite{masry2024chartinstruct} & Swin Transformer \cite{liu2021swin} & Flan-T5-XL's encoder \cite{chung2022flant5} & Flan-T5-XL's decoder \cite{chung2022flant5} & 3B \\
        StructChart \cite{xia2023structchart} & $\text{ViT-B}_{\text{16}}$ \cite{dosovitskiy2020image} & - & Transformer \cite{DBLP:conf/nips/Transformer17} & 282M \\
        
        \midrule
        \multicolumn{5}{c}{\textit{Large Vision-language Models}} \\
        \midrule
        ChartInstruct-Llama \cite{masry2024chartinstruct} & Swin Transformer \cite{liu2021swin} & - & Llama2-7B \cite{touvron2023llama2} & 7B \\
        MMCA \cite{liu2023mmc} &  $\text{CLIP ViT-L}_{\text{14}}$ \cite{radford2023clip} & - & Llama-7B \cite{touvron2023llama} & 7B \\
        ChartLlama \cite{han2023chartllama} & $\text{CLIP ViT-L}_{\text{14}}$ \cite{radford2023clip}  & - & Vicuna-13B \cite{chiang2023vicuna} & 13B \\
        ChartAssistant-S \cite{meng2024chartassisstant} & $\text{CLIP ViT-L}_{\text{14}}$ \cite{radford2023clip} & - &  Llama-13B \cite{touvron2023llama} & 13B \\
        ChartVLM \cite{xia-etal-2024-chartx}  & $\text{ViT-B}_{\text{16}}$/$\text{ViT-L}_{\text{16}}$ \cite{dosovitskiy2020image} & - & Vicuna-13B \cite{chiang2023vicuna} & 13B \\

        \bottomrule
    \end{tabular}
    }
    \end{adjustbox}
    \vspace{-5mm}
    
    \label{tab:model_summary}
    
\end{table*}

\subsubsection{Large Vision-language Models}

While the aforementioned pre-trained approaches significantly enhance chart understanding, they require task-specific fine-tuning (see \Cref{fig:pretrained_vs_lvlms}), limiting their generalizability. Contrastingly, large foundation models have demonstrated strong zero-shot or few-shot capabilities. For instance, in text generation, LLMs like GPT-3 \cite{brown2020language} showcase remarkable generalization to novel tasks when the model size and pre-training data reach certain scales. These models are able to handle new tasks in a zero-shot or few-shot manner, obviating the need for model retraining. Extending this paradigm into image understanding, studies have stacked vision encoders on top of LLMs, including methods such as LLaVA \cite{liu2023improvedllava} and mPLUG-Owl \cite{ye2023mplug}, significantly advancing this field. These larger vision-language models (LVLMs) differ from the smaller vision-language models introduced previously by having substantially greater model sizes and instruction tuning. Unique to these LVLMs is the inclusion of a visual-language projector, $\textsc{Projector}_{\text{vt}}$, given the vision encoders and LLMs' pre-training scales. Here, the visual representations obtained from the vision encoder are fed to $\textsc{Projector}_{\text{vt}}$ prior to concatenated with text representations.

Building upon general-purpose LVLMs \cite{liu2023improvedllava, ye2023mplug, lin2023sphinx}, a couple of very recent concurrent studies proposed instruction-tuning datasets and models designed for chart understanding: MMCA \cite{liu2023mmc}, ChartLlama \cite{han2023chartllama}, ChartAssistant \cite{meng2024chartassisstant}, and ChartInstruct \cite{masry2024chartinstruct}. By continuing training these general-purpose LVLMs on tailored instruction-tuning datasets, these models have profoundly expanded the capabilities of LVLMs in chart understanding. Below, we will discuss their difference in generating instruction-following datasets and training LVLMs. 

Regarding instruction-tuning approaches, the four LVLMs leverage distinct strategies. MMCA and ChartInstruct employ a two-stage training pipeline, a technique also observed in several LVLMs \cite{ye2023mplug,chen2023shikra,liu2023improvedllava}. The initial stage focuses on aligning the visual and textual representations, during which only the projector is trainable, and the other components are frozen. Subsequently, in the second stage, both the projector and the text decoder undergo training, while the visual encoder remains unchanged. In contrast, ChartLlama and ChartAssistant adopt a one-stage instruction-tuning method, eliminating the isolated projector training. A notable distinction in their methodologies is ChartAssistant undergoing an extra chart-to-table pre-training process. The two-stage training method introduces complexity and cost over single-stage approaches. Furthermore, the tangible benefits of isolating the projector during training on overall model performance remain to be conclusively established. \looseness=-1

The datasets utilized for instruction tuning vary among these models, as outlined in \Cref{tab:it_dataset_summary}. MMCA's dataset encompasses the broadest range of tasks, while ChartAssistant's dataset is the largest. These datasets are a mix of real-world and synthetic charts, sourced from existing chart collections or the internet. On the other hand, ChartLlama generates data and charts completely from scratch. Both ChartAssistant and ChartLlama leverage visualization tools like Matplotlib to create charts with diverse styles. ChartInstruct is the only dataset that is entirely composed of real-world charts.

\begin{table*}[t]
    \small
    \centering
    \caption{Summary of pre-training and instruction-tuning datasets for chart understanding. \cmark~/\xmark~ indicates the dataset consists of both types of data or charts.}%
    \begin{adjustbox}{max width=0.95\textwidth}
    {
    \begin{tabular}{lrrcccc}
        \toprule
        \textbf{Model} & \textbf{\# Instances} & \textbf{\# Tasks}  & \textbf{Real-world Data} & \textbf{Real-world Charts} & \textbf{Synthetic Data Augmentation}\\
        
        \midrule
        \multicolumn{6}{c}{\textit{Pre-training Datasets}} \\
        \midrule
        ChartT5 \cite{zhou-etal-2023-enhanced} & 495K & 2 &\cmark~/\xmark~ & \cmark~/\xmark~ & \xmark~ \\
        MatCha \cite{liu2023mmc} & 46M & 5 &\cmark~/\xmark~ & \cmark~/\xmark~ & \xmark~ \\
        UniChart \cite{liu2023mmc} & 7M & 4 &\cmark~/\xmark~ & \cmark~/\xmark~ & \cmark~ \\
        \midrule
        \multicolumn{6}{c}{\textit{Instruction-tuning Datasets}} \\
        \midrule
        MMCA \cite{liu2023mmc} & 2M & 9 &\cmark~ & \cmark~/\xmark~ & \xmark~ \\
        ChartLlama \cite{han2023chartllama}  & 160K & 7  & \xmark~ & \xmark~ & \cmark~\\
        ChartAssistant \cite{meng2024chartassisstant} & 39M  & 5 &  \cmark~ & \cmark~/\xmark~ & \cmark~\\
        ChartInstruct \cite{masry2024chartinstruct} & 191K & 9 & \cmark~ & \cmark~ & \xmark~ \\ 
        \bottomrule
    \end{tabular}
    }
    \end{adjustbox}
    \vspace{-5mm}
    
    \label{tab:it_dataset_summary}
    
\end{table*}

Direct comparisons among existing LVLMs designed for chart understanding are challenging due to the absence of uniform evaluation standards. Although ChartAssistant surpasses others on the several chart understanding benchmarks (see \Cref{tab:sota_performance}), it remains uncertain which specific factors (e.g. model architecture, training data, training scheme) drive its superior performance. A more thorough experimental framework is needed to dissect how various factors enhance a model's chart comprehension capabilities. Future research can explore several pertinent questions discussed in the following paragraphs.  \looseness=-1

First, \textbf{how do different visual representations affect an LVLM's chart understanding proficiency?} Examining how different visual representations, such as those trained with constrastive learning objectives versus the alternatives, impact an LVLM's capacity to comprehend and generate accurate descriptions or answers based on charts could reveal insights into pre-training strategies. Additionally, all LVLMs shown in \Cref{tab:model_summary} use $\text{CLIP ViT-L}_{\text{14}}$, a general-purpose vision encoder. Conducting experiments by swapping this vision encoder with other vision encoders for charts, such as those from UniChart and MatCha, before instruction tuning can reveal the importance of a vision encoder tailored for chart understanding. Furthermore, some studies have found that LVLMs with a vision encoder based on image patches are ill-suited to solve geometric problems \cite{lu2023mathvista, yue2023mmmu}, tasks that are highly related to chart understanding. This could be an explanation for the limited chart captioning abilities of state-of-the-art LVLMs, including GPT-4V \cite{huang2023lvlms}. A deep dive into this problem could help researchers design suitable image pre-processing methods to overcome this limitation. %

Second, \textbf{should we use base (i.e. non-instruction-tuned) language models or those that have undergone instruction tuning?} Many of the existing LVLMs use an instruction-tuned LLM as the text decoder \cite{liu2023improvedllava, ye2023mplug, liu2023mmc}. Since the LLMs have already been trained on text-only instruction-tuning datasets, the resulting LVLMs may be more capable of following instructions. Given that these LLMs have been pre-trained on instruction-tuning datasets, which include \textit{anti-harmfulness} instances against toxic or inappropriate content, the LVLMs derived from them are naturally better at following instructions while mitigating harmful outputs. However, the potential drawbacks of instruction tuning, such as introducing bias or diminishing performance, warrant exploration to discern the trade-off between instructions following, safeguarding against harmful content, and maintaining performance.

Third, \textbf{is there a necessity for a multi-stage training process?} This question probes whether the benefits provided by multi-stage training, potentially yielding more nuanced and sophisticated representations, justifing its added complexity and cost. The performance of ChartLlama, which employs a one-stage training approach, outperforming MMCA, which uses a two-stage pipeline, with less instruction-tuning data, suggests multi-stage training may not be indispensable. Nevertheless, differences in instruction-tuning data quality, model architecture, and training configurations between MMCA and ChartLlama underscore the importance of controlled experiments to evaluate training paradigms rigorously. \looseness=-1

Finally, \textbf{is it possible for models trained exclusively on synthetically generated data to match or exceed the performance of those trained on datasets curated from human-generated content?} While synthetic data offers the advantage of scalability and control over diversity and complexity, it may lack the nuanced and unpredictable variations present in human-generated charts. Exploring whether synthetic data can truly capture the breadth of human creativity and intricacy in chart design is essential, as is understanding the potential gaps that might arise when such models are applied to real-world data. Recent work \cite{masry2024chartinstruct} has shown that completely removing real-world data from the training set significantly reduces performance. However, adding synthetic training data helps improve data efficiency and reduces the need for large-scale real-world chart data. Future research can focus on studying the effectiveness of synthetic data at varying scale and on developing methods for creating synthetic data that can better mimic real-world complexities. \looseness=-1

Together, these questions highlight a comprehensive spectrum of research avenues endeavoring to decode the intricate processes underlying chart comprehension through LVLMs. A detailed investigation of these topics promises to advance our progress towards models that can interpret an extensive array of charts with a level of proficiency comparable to or surpassing human capability.

\subsection{Tool Augmentation}

\label{subsec:tool_aug}

Tool augmentation refers to utilizing external systems to address limitations in modeling capabilities, such as restricted visual representations \cite{qin2023tool}. Within the domain of chart understanding, these augmentation tools primarily focus on extracting key information from charts to facilitate further processing by more advanced models. Early-adopted tools are OCR systems designed to extract textual content from images. These systems are categorically divided into those that are enriched by language semantics \cite{shi2016robust,cheng2017focusing,bautista2022scene,fang2022abinet++}, and those that do not incorporate language semantics \cite{shi2016end,wang2017gated,liu2016star,borisyuk2018rosetta}. Despite their utility, the quality of information extracted using traditional OCR systems for chart understanding remains suboptimal due to their overall performance challenges. Hence, recent advancements have shifted towards OCR-free, end-to-end pre-trained vision-language models, such as Donut \cite{kim2022donut} and Pix2Struct \cite{lee2023pix2struct}. %
\begin{table*}[t]

    \small
    \centering
    \caption{The state-of-the-art performance on different tasks and datasets. We only show datasets where modern approaches based on pre-trained vision-language models or LVLMs have benchmarked due to the limitations of older datasets. Note that for long-form chart understanding tasks (i.e. Open-domain Chart Question Answering, Chart Captioning, and Chart Caption Factual Error Correction), conventional metrics, such as BLEU, are sub-optimal and do not truly reflect models' performance on these tasks, as discussed in \Cref{subsec:metrics}. Despite advancements in metrics such as ChartVE, human evaluation remains essential for a thorough assessment of chart understanding performance.} 
    \begin{adjustbox}{max width=0.98\textwidth}
    {
    \begin{tabular}{lllccc}
        \toprule
        \textbf{Task} & \textbf{Dataset} & \textbf{SOTA Model} & \textbf{Fine-tuning} & \textbf{Performance (\%)} & \textbf{Metric} \\
        \midrule
        \multirow{3}{*}{Chart-to-Table Conversion} & PlotQA & DePlot \cite{liu-etal-2023-deplot} & \cmark~ & 94.2 & $\texttt{RMS}_{\text{F1}}$  \\
                                                   & ChartQA & ChartAssistant-D \cite{meng2024chartassisstant} & \cmark~ & 92.0 & $\texttt{RMS}_{\text{F1}}$ \\
                                                   & VisText & Chart-to-Tabe \cite{huang2023lvlms} & \cmark~ & 83.6 & $\texttt{RMS}_{\text{F1}}$\\

        \midrule
        \multirow{2}{*}{Factoid Chart Question Answering}  & ChartQA & InternVL 1.5 \cite{chen2024far} & \xmark~ & 83.8 & \texttt{RA} \\
                                                   & ChartBench & GPT-4V \cite{openai2023gpt4v} & \xmark~ & 54.4  & \texttt{Acc+} \\

        \midrule
        \multirow{2}{*}{Chart Fact-checking}   & ChartFC & ChartInstruct-Flan-T5-XL \cite{masry2024chartinstruct} & \cmark~ & 72.7 & Accuracy \\
                                        & ChartCheck & DePlot-DeBERTa \cite{akhtar2023chartcheck} & \cmark~ & 73.8 & Accuracy \\

        \midrule
        \multirow{3}{*}{Factual Inconsistency Detection for Chart Captioning} & $\text{Chocolate}_{\text{FT}}$ & Bard \cite{team2023gemini} & \xmark~ & 29.1 & Kendall's Tau \\
                                                                              & $\text{Chocolate}_{\text{LLM}}$ & GPT-4V \cite{openai2023gpt4v} & \xmark~  & 20.5 & Kendall's Tau \\
                                                                              & $\text{Chocolate}_{\text{LVLM}}$ & ChartVE \cite{huang2023lvlms} & \cmark~  & 17.8 & Kendall's Tau \\
        
        \midrule
        Open-domain Chart Question Answering &  OpenCQA & ChartAssistant-S \cite{meng2024chartassisstant} & \xmark~ & 20.2 & BLEU \\
        \midrule
        
        \multirow{2}{*}{Chart Captioning} & $\text{Chart-to-Text}_{\text{Statista}}$ & ChartAssistant-S \cite{meng2024chartassisstant} & \cmark~  &  41.0 & BLEU \\
                                          & $\text{Chart-to-Text}_{\text{Pew}}$ & ChartAssistant-S \cite{meng2024chartassisstant} & \cmark~ &  16.5 & BLEU \\

        \midrule 
        \multirow{3}{*}{Chart Caption Factual Error Correction}   & $\text{Chocolate}_{\text{FT}}$ & C2TFEC \cite{huang2023lvlms} & \xmark~ & 81.1 & Bard \\
                                                                              & $\text{Chocolate}_{\text{LLM}}$ & GPT-4V \cite{openai2023gpt4v} & \xmark~ & 52.4 & GPT-4V \\
                                                                              & $\text{Chocolate}_{\text{LVLM}}$ & C2TFEC \cite{huang2023lvlms} & \xmark~ & 34.3 & ChartVE \\
        
        \bottomrule
    \end{tabular}
    }
    \end{adjustbox}

    \label{tab:sota_performance}
    
\end{table*}

Parallel to OCR-free vision-language models, another line of work focuses on harnessing the strong reasoning abilities of LLMs and develops tools to augment LLMs with extracted information from charts. These tools are either specialized in converting charts to the underlying data tables, such as DePlot \cite{liu-etal-2023-deplot}, StructChart \cite{xia2023structchart}, and Chart-to-Table \cite{huang2023lvlms}, or are equipped with additional capabilities for chart question answering, exemplified by DOMINO \cite{wang2023domino}. Due to these models' proficiency in extracting chart information, they enable a strategic decompose the two required abilities of chart understanding: \textit{perception} and \textit{reasoning}. \textit{Perception} involves extracting information from charts, such as data tables or specific values, whereas \textit{reasoning} entails text-based logical and mathematical operations over the extracted data. When the extracted information is linearized into text sequences, it becomes feasible for LLMs to employ their strong reasoning abilities. Compared with LVLMs, tool-augmented LLMs have demonstrated superior performance in interpreting various chart components, such as labels, values, and trend lines \cite{huang2023lvlms}. However, a notable limitation is their inability to interpret visual attributes within charts, such as colors and the spatial arrangement of data points, since these details are typically absent in the extracted information.

To mitigate this limitation, a potential solution might involve augmenting the data tables to capture color and spatial information \cite{do2023llms}. Nevertheless, this solution may fall short in cases involving unique chart types or when the color schemes of different data points are semantically similar (e.g. blueish-green vs. greenish-blue). Thus, while tool-augmented LLMs exhibit commendable performance in tasks that prioritize textual and numerical understanding within charts over visual attribute interpretation, such as chart captioning, applicability remains limited in scenarios necessitating a detailed understanding of visual attributes.

\subsection{State-of-the-art Performance}

In \Cref{tab:sota_performance}, we display the state-of-the-art performances across various tasks and datasets. The best performance is often achieved by pre-trained models and LVLMs. This highlights the profound impact of leveraging larger-scale vision and language models, coupled with pre-training or instruction-tuning techniques, in advancing the capabilities of chart understanding models. Additionally, it can be observed that chart understanding tasks that involve long-form text generation (i.e. Chart Captioning) are more challenging than those with gold-standard ground truth answers (e.g. Chart Question Answering). To further improve the performance on these difficult tasks, future work can study the research questions outlined in \Cref{subsec:generation_based} and \Cref{subsec:tool_aug}.

\section{Challenges and Future Directions} \label{Sec:5}

\subsection{Domain-specific Charts}
\label{subsec:domain_specific_charts}

In \Cref{subsec:dataset_source}, we have highlighted the predominance of datasets that are not tailored to any specific domain and have recognized the scarcity of domain-specific chart understanding datasets. Although a few datasets like PaperQA \cite{lu2023mathvista} and SciCap \cite{hsu-etal-2021-scicap-generating} pave the way to exploring charts in scientific literature, the majority of chart types featured therein, such as \textit{bar}, \textit{line}, and \textit{pie}, remain commonplace. Expanding the dataset spectrum to encompass unique chart formats prevalent in specialized disciplines would illuminate struggles that contemporary models might currently overlook.

The chemistry domain serves as an example in which automatic chart understanding can bring significant benefit for downstream applications, such as drug repurposing report generation \cite{wang2021covid19} and drug discovery \cite{edwards2023synergpt}. Charts in the chemistry domain, like pathway flowcharts, nuclear magnetic resonance (NMR) spectra, and molecular orbital diagrams, pose distinctive challenges. For instance, pathway flowcharts amalgamate a diverse range of data types, from quantitative measurements of enzyme activity to qualitative representations of interactions, necessitating the recognition of graphic components and the comprehension of their interconnectedness. Interpreting NMR spectra demands an analysis of peak patterns and intensities, alongside the molecule's inferred structural attributes, which goes beyond mere data retrieval to encompass complex interpretative reasoning predicated on chemical knowledge. Similarly, molecular orbital diagrams represent a conceptual visualization of quantum mechanics, requiring deep insight into the complexities of atomic and molecular orbital interactions.

The specialized charts found in domains such as chemistry often incorporate technical jargon and unique symbols intrinsic to their fields, presenting interpretive hurdles that existing chart understanding models may struggle to clear without the infusion of specific domain expertise. In developing domain-specific datasets, several pertinent research questions arise. First, do current approaches based on large-scale pre-training on datasets with general chart types enable models to generalize to unseen domain-specific charts? Second, how instrumental is domain expertise in enhancing the comprehension abilities for such charts? And third, if domain knowledge is helpful, what are the most effective methods for integrating such expertise into chart understanding models?

\subsection{Evaluation}
\label{subsec:eval}
As highlighted in \Cref{subsec:dataset_source}, prior work on long-form chart understanding tasks, including open-domain chart question answering, chart captioning, and chart caption factual error correction, only utilized conventional lexical-based metrics to assess output quality, which is inadequate. We propose five pivotal criteria for a more precise evaluation: \textit{faithfulness}, \textit{coverage}, \textit{relevancy}, \textit{robustness}, \textit{fairness} and \textit{bias}. %

\textbf{Faithfulness.} The faithfulness of a model's output refers to the extent to which the information presented is accurate and consistent with the data depicted in the chart. Unfaithfulness or factual inconsistency can lead to outputs that misinform users \cite{huang-etal-2023-zero, chan-etal-2023-interpretable, kim-etal-2024-can}, potentially causing misinterpretations or incorrect decision-making based on inaccurate chart understanding. This is of particular significance in domains where accurate data representation is critical, such as education and news reporting. Although the \textsc{ChartVE} metric \cite{huang2023lvlms} offers a robust assessment of the factual consistency between chart data and textual outputs, it is unable to perform granular evaluations that locate errors within text spans. Future studies can emphasize the development of finer-grained evaluation techniques alongside the classification of error types, such as the value errors and trend errors delineated by previous work \cite{huang2023lvlms}, or the confidence boundary of model output hallucinations \cite{zhang2023rtuning}. Such approaches would provide a clearer picture of where models are failing in terms of faithfulness.  \looseness=-1

\textbf{Coverage.} The notion of coverage analyzes whether a model's output encompasses all the essential insights that a chart conveys. Models with lower coverage might overlook key elements or trends within the chart, providing a partial or skewed summary of the data. This is particularly critical when comparing fine-tuned task-specific models to large vision-language models \cite{huang2023lvlms}. Existing evaluation metrics falter in their ability to measure coverage effectively. To truly gauge the breadth of information captured by a model from a chart, evaluation metrics that are capable of quantifying coverage are needed. These should account for the presence, significance, and the instrumental contribution of key insights toward an integrated comprehension of the chart. \looseness=-1

\textbf{Relevancy.} Relevancy is a central concern for open-domain chart question answering tasks. Beyond faithfulness and coverage, it is vital to evaluate if the model's output pertains specifically to the question posited. The importance of relevancy stems from the possibility that an output, while being faithful to the chart, may only tangentially relate to the user's question. This mismatch can result in outputs that, although factually correct, are of limited utility to the user. Assessing relevancy necessitates a nuanced understanding of the user's intent and the context of the question, ensuring that responses are not only accurate and comprehensive but also specifically tailored to fulfill the user's informational needs. \looseness=-1

\textbf{Robustness.} Robustness is a crucial aspect of interaction with LVLMs, particularly when using prompts as open-ended questions to generate descriptions (answers) from input charts. The issue arises when models produce significantly varied outputs in response to subtly different but semantically similar prompts. Such variability can be problematic, as users expect consistent responses to similar queries and safeguards against adversarial prompts \cite{li-etal-2023-defining}. To evaluate a model's generation quality effectively, it is vital to assess its performance across a range of prompts that pertain to the same question or topic. Drawing inspirations from previous research in evaluating the robustness of LLMs \cite{sclar2023quantifying}, future studies can design adversarial prompts that challenge the model's understanding of charts. This approach helps in determining the model's robustness and reliability in interpreting and generating accurate descriptions under varied prompting conditions.

\textbf{Fairness and Bias.} 
Evaluating fairness and mitigating biases in the generation of models' content is critical when deploying these tools in user-facing applications. There are two main forms of bias to be aware of: \textit{fairness bias}, which concerns the equitable representation and treatment of various groups (\textit{e.g.,} race, culture, or gender) \cite{qiu-etal-2023-gender,Yang_Yu_Fung_Li_Ji_2023}, and \textit{sample bias}, which occurs when the data used to train the model is not representative of the broader population or scenario it's intended to serve (\textit{e.g.}, a line chart dataset containing only monotonic trends). To ensure outputs encompass diverse perspectives, it is crucial to identify and rectify these biases. Fairness bias often emerges when models disproportionately rely on their pre-trained knowledge, possibly neglecting the specific visual cues present in a chart. This reliance can lead to outputs that do not accurately reflect the visual data. For instance, if a model is trained on a dataset in which the charts consistently depict a particular group as having a higher diabetes rate compared to others, it is highly probable that when this model encounters a chart presenting contradictory information, it may still generate descriptions that align with its training, favoring interpretations that match the patterns it has learned during pre-training, even if those interpretations are less likely given the new data. Meanwhile, sample bias is evident when datasets over-represent certain types of charts. For example, some models, such as MatCha, exhibit proficiency in interpreting line charts where data points follow a monotonic sequence, but struggle to comprehend line charts characterized by convex or concave configurations, as well as those with fluctuating trends \cite{huang2023lvlms}. To address these issues, future work can develop benchmarks to include charts designed to counteract these biases and feature complex configurations, deviating from the typical charts in existing datasets. Future strategies should focus on diversifying training data and using adversarial training techniques to mitigate these biases. This will promote a more equitable and inclusive approach to content generation. %

One prominent direction for evaluation is model-based metrics. As evidenced in \Cref{tab:sota_performance}, task-specific models like ChartVE and general-purpose models like GPT-4V achieve great correlation with human judgment in terms of faithfulness on the Chocolate dataset. This suggests that leveraging models themselves as evaluators offers a promising path forward in assessing chart understanding outputs, particularly in dimensions such as faithfulness where traditional lexical metrics fall short. Moving beyond the reliance on human-annotated gold standards, model-based evaluation can provide scalable, consistent, and nuanced insights into chart understanding performance across various dimensions. This approach, however, necessitates the development of reliable and unbiased model evaluators, which may entail pre-training models specifically for the purpose of evaluation or fine-tuning existing models with carefully curated evaluation data. Future work could explore the creation of meta-models trained on datasets designed to capture a wide array of evaluation criteria, thereby offering deeper and more comprehensive assessments of chart understanding capabilities. This approach to evaluation could significantly guide researchers towards targeted improvements in various chart understanding tasks.

\subsection{Dissecting LVLM Components}
In \Cref{subsec:generation_based}, we delved into the design choices of LVLMs and underscored the myriad challenges they entail, alongside potential future directions for research. A pivotal area of inquiry lies in the impact of various visual representations on the chart understanding proficiency of LVLMs. The question of whether differing visual encoding methods, such as those learned through contrastive objectives versus conventional approaches, affect an LVLM's ability to accurately interpret and generate insights from charts forms a crucial consideration. Moreover, the exploration of incorporating fused visual representations to potentially enhance model interpretability signals a promising avenue for enhancing data comprehension.

Furthermore, the decision between utilizing base language models versus those that have undergone instruction tuning also emerges as a key point of discussion. The nuanced trade-offs between instruction-following capabilities, the safeguarding against harmful outputs offered by instruction-tuned models, and the potential performance impact underscore the need for a balanced approach in leveraging these advanced language models for chart understanding tasks. 

Lastly, the inquiry into the necessity of a multi-stage training process versus the sufficiency of models trained exclusively on synthetically generated data presents a compelling debate. The potential of synthetically generated datasets to adequately capture the complexity and nuance of human-generated content remains an open question. This encompasses not only the generalization capabilities of models trained on such datasets but also the potential for biases or limitations induced by synthetic data generation methodologies. These challenges highlight the importance of meticulous dataset curation, model training strategies, and the integration of domain-specific knowledge to overcome the inherent limitations of current LVLMs in chart understanding applications.

\subsection{Agentic Settings} %
AI agents are artificial entities that sense their environment, make decisions, and take actions. Recent studies have explored the feasibility of using large models as agents for applications such as web browsing \cite{deng2023mindweb}, house-holding \cite{liu2024agentbench}, and scientific problem-solving \cite{yang2024llm}. However, no prior work has studied LVLMs as chart interpretation agents. One promising agentic application of LVLMs is education. In teaching environments, these models could assist in breaking down complex charts, explaining them in simpler terms, and even generating quizzes or interactive activities based on the chart data. For students struggling with data literacy, having an intelligent agent that can tailor explanations to their level of understanding and link chart data to real-world applications could foster a deeper comprehension of subjects. Such educational applications signify a step towards more personalized and accessible learning experiences, leveraging the power of LVLMs to demystify complex information for learners of all levels.

Additionally, report generation can also benefit significantly from LVLMs as chart interpretation agents. In various industries, from finance to healthcare, the ability of an AI agent to automatically search for relevant chart data within internal databases and generate reports that highlight critical data points as well as recommending strategic actions could significantly enhance efficiency and decision-making processes. More importantly, report generation could also involve providing visual insights by synthesizing new charts via visualization tools based on the information the agents gathered. This could be particularly help in scenarios where rapid data interpretation is crucial, such as in stock market analysis or emergency medical response planning.

\subsection{Multilingual Chart Understanding}

Despite significant advances in chart understanding technologies, the aspect of multilingual chart understanding remains underexplored. Much of the current research and dataset development focuses predominantly on English, limiting the applicability of these models in a global context. The need for multilingual chart understanding arises from the global nature of data dissemination, where charts are created and shared in diverse languages, catering to a broad audience. A couple of research directions arise to understand models' ability to generate insights from charts across languages. First, while current state-of-the-art LVLMs, GPT-4V, demonstrate satisfactory performance in understanding charts \cite{lu2023mathvista}, particularly when the values are labeled next to the corresponding data points \cite{huang2023lvlms}, does this hold true for charts in other languages as well? Second, what if the texts within the charts and the textual queries are in different languages? Are LVLMs capable of cross-lingual understanding in such cross-modality settings? Finally, investigating the potential for LVLMs to adapt to multilingual contexts without extensive retraining or additional instruction-tuning presents an intriguing avenue for future research. The capability of models to seamlessly transition across languages, interpreting and generating insights with high fidelity, could significantly broaden the accessibility and usability of chart understanding technologies globally.

\section{Conclusion}
In this survey, we provide a thorough examination of the evolving field of automatic chart understanding, with an emphasis on recent developments such as the transformative impact of large vision-language models. Our review starts with a detailed discussion on the diverse datasets that drive chart understanding research, analyzing their sources, diversity, and the specific tasks they support. We then categorizes the landscape of chart understanding, covering from foundational classification-based and generation-based models to the cutting-edge applications of LVLMs. We offer an in-depth overview of methodologies. Additionally, we critically examine the role of datasets and outline future research trajectories. Our survey aims to spur innovation and exploration in the vibrant field of chart understanding. This marks a pathway for future breakthroughs in this crucial area of study.

\bibliographystyle{IEEEtran}
\bibliography{main.bib}

\vspace{-50pt}
\begin{IEEEbiography}[{\includegraphics[width=1in,
height=1.25in,clip,keepaspectratio]{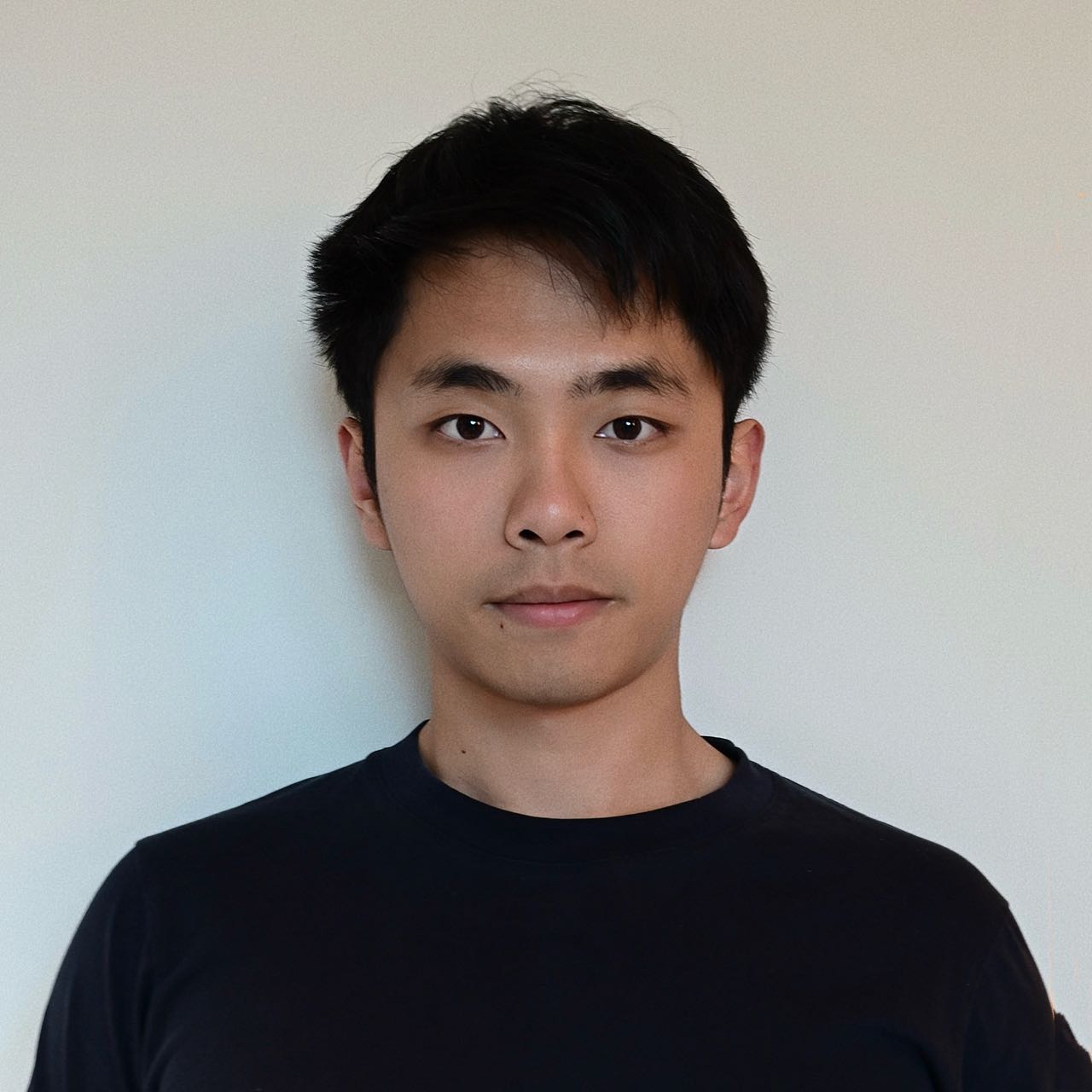}}]{Kung-Hsiang Huang} is a research scientist at Salesforce Research. He obtained his  Ph.D. from the University of Illinois Urbana-Champaign under the guidance of Prof. Heng Ji. His research focuses on fact-checking, faithfulness enhancement, factual error correction, and chart understanding. He is a recipient of the inaugural fellowship awarded by the Amazon-Illinois Center. He has taught two tutorials at conferences, including ``KDD'22: The Battlefront of Combating Misinformation and Coping with Media Bias''. He is the recipient of the Top Reviewer Award at NeurIPS 2024.%
\end{IEEEbiography}
\vspace{-40pt}
\begin{IEEEbiography}[{\includegraphics[trim= 0 0.3in 0 0.5in, width=1in,height=1in,clip,keepaspectratio]{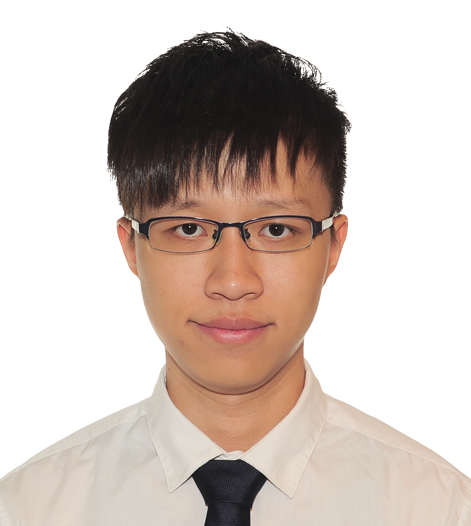}}]{Hou Pong Chan} is a research scientist at the Language Technology Lab, Alibaba DAMO Academy. Prior to that, he was a visiting postdoc researcher at the University of Illinois at Urbana-Champaign and was a visiting lecturer (Macao Fellow) at the University of Macau. He received a Ph.D. degree in Computer Science from The Chinese University of Hong Kong. His research focuses on improving the factuality and controllability of natural language generation. He has published more than 20 research papers in leading NLP conferences (e.g., ACL, EMNLP, and NAACL) and journals (e.g., TACL). He is a recipient of the Outstanding Reviewer Award at ACL 2021 and EMNLP 2020. %
\end{IEEEbiography}
\vspace{-35pt}
\begin{IEEEbiography}[{\includegraphics[trim= 0 0 0 0.3in, width=1in,height=1in,clip,keepaspectratio]{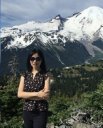}}]{Yi R.(May) Fung} is a Tenure-Track Assistant Professor at the Hong Kong University of Science and Technology Department of Computer Science and Engineering. She completed her PhD in computer science at the University of Illinois, with Prof. Heng Ji, after which she spent time visiting MIT as a postdoctoral affiliated researcher at Prof. Paul Liang's lab. May leads research on multimedia knowledge reasoning, misinformation detection, and computation for the social good with human-centered, scalable oversight principles. May is a recipient of the ACL'24 Outstanding Paper Award, NAACL'24 Outstanding Paper Award, NAACL'21 Best Demo Paper, the UIUC Lauslen and Andrew fellowship, and the UIUC Yunni \& Maxine Pao Memorial Fellowship. She has also been previously selected for invited talks at the Harvard Medical School Bioinformatics Seminar and the Hong Kong University Junior Scholar Seminar. She has taught two tutorials at conferences, including \textit{KDD'22: The Battlefront of Combating Misinformation and Coping with Media Bias}. %
\end{IEEEbiography}
\vspace{-50pt}
\begin{IEEEbiography}[{\includegraphics[width=1in,height=1in,clip,keepaspectratio]{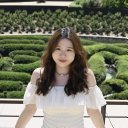}}]{Haoyi Qiu} is a first-year PhD student in Computer Science at UCLA advised by Prof. Nanyun (Violet) Peng. Prior to that, she graduated from the University of Michigan, with a B.S. in Computer Science and Math, advised by Prof. Joyce Y. Chai. Her research focuses on making AI more trustworthy and improve its factuality and safety. She is a recipient of the Outstanding Reviewer Award at EMNLP 2023. %
\end{IEEEbiography}
\vspace{-30pt}
\begin{IEEEbiography}[{\includegraphics[trim= 0 0.4in 0 0in, width=1in,height=1in,clip,keepaspectratio]{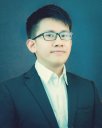}}]{Mingyang Zhou}is an applied researcher at Capital One. Previously, he was a post doctoral research scientist a the Department of Electrical Engineering, Columbia University and he obtained his Ph.D degree from University of California, Davis. His research interest lies in vision and language pre-training and visual dialogue system. His recent works include the ChartT5 chart understanding model, and the first work on Unsupervised Vision-and-Language Pre-training via Retrieval-based Multi-Granular Alignment. He is a winner of Alexa Social Bot Challenge in 2018. %
\end{IEEEbiography}
\vspace{-30pt}
\begin{IEEEbiography}[{\includegraphics[width=1in,height=1in,clip,keepaspectratio]{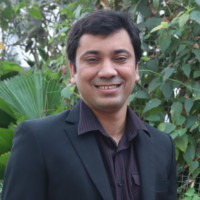}}]{Shafiq Joty} is a research director at Salesforce Research and an Associate Professor at the Computer Science Department of the Nanyang Technological University. His research has contributed to over 30+ patents and 140+ papers in top-tier NLP and ML conferences and journals. He has served as the Program Chair of SIGDIAL'23, a member of the best paper award committees for ICLR'23 and NAACL'22, and in the capacity of a (senior) area chair for many leading NLP and ML conferences (e.g. NeurIPS, EMNLP, and ACL). He previously gave tutorials at EMNLP'23, IEEEVis'22, ACL'19,  COLING'18 and ICDM'18. %
\end{IEEEbiography}
\vspace{-30pt}
\begin{IEEEbiography}[{\includegraphics[width=1in,height=1in,clip,keepaspectratio]{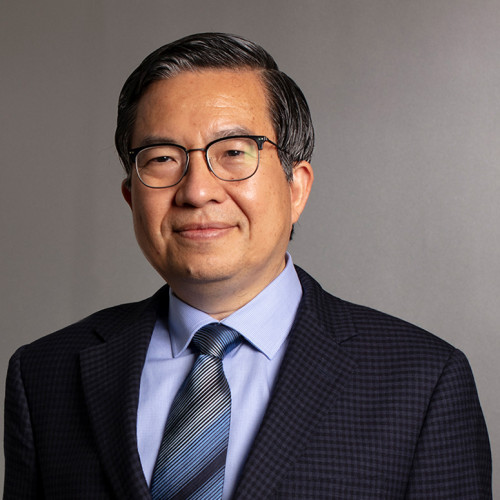}}]{Shih-Fu Chang} is the Dean of Columbia Engineering and the Morris A. and Alma Schapiro Professor in the Electrical Engineering Department and Computer Science Department. A primary goal of his work is to develop intelligent systems that can extract rich information and knowledge from diverse data of multiple modalities, including images, video, language, and audio. His scholarly impacts can be seen in peer-reviewed publications, best paper awards, 30+ issued patents, and technology transfer. For his long-term contributions, he was awarded the IEEE Signal Processing Society Technical Achievement Award, ACM SIGMM Technical Achievement Award, the Honorary Doctorate from the University of Amsterdam, and the IEEE Kiyo Tomiyasu Award. He received the Great Teacher Award from the Society of Columbia Graduates. He served as Chair of ACM SIGMM, Chair of Columbia Electrical Engineering Department. He is a Fellow of the American Association for the Advancement of Science (AAAS), ACM, and IEEE, and an elected Academician of Academia Sinica. 
\end{IEEEbiography}
\vspace{-30pt}
\begin{IEEEbiography}
[{\includegraphics[width=1in,height=1in,clip,keepaspectratio]{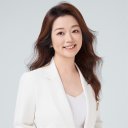}}]{Heng Ji} is a professor at Computer Science Department, and an affiliated faculty member at Electrical and Computer Engineering Department and Coordinated Science Laboratory of University of Illinois Urbana-Champaign. She is an Amazon Scholar and the Founding Director of Amazon-Illinois Center on AI for Interactive Conversational Experiences. Her research interests focus on Natural Language Processing, especially on Multimedia Multilingual Information Extraction, Knowledge-enhanced Large Language Models, Knowledge-driven Generation, and Conversational AI. She was selected as ``Young Scientist'' by the World Economic Forum in 2016 and 2017 and was named as part of Women Leaders of Conversational AI (Class of 2023) by Project Voice. The awards she received include ``AI's 10 to Watch'' Award by IEEE Intelligent Systems in 2013, NSF CAREER award in 2009, PACLIC2012 Best paper runner-up, ``Best of ICDM2013'' paper award, ``Best of SDM2013'' paper award, ACL2020 Best Demo Paper Award, NAACL2021 Best Demo Paper Award, Google Research Award in 2009 and 2014, IBM Watson Faculty Award in 2012 and 2014 and Bosch Research Award in 2014-2018. 
\end{IEEEbiography}
\vspace{-40pt}

\vfill

\end{document}